%% file: main.tex
\documentclass[letterpaper]{article} 
\usepackage{aaai2026}  
\usepackage{times}  
\usepackage{helvet}  
\usepackage{courier}  
\usepackage[hyphens]{url}  
\usepackage{graphicx} 
\usepackage{natbib}  
\usepackage{caption} 
\usepackage{algorithm}
\usepackage{algpseudocode}

\usepackage{newfloat}
\usepackage{listings}
\DeclareCaptionStyle{ruled}{labelfont=normalfont,labelsep=colon,strut=off} 
\floatstyle{ruled}
\newfloat{listing}{tb}{lst}{}
\floatname{listing}{Listing}
\title{Assessing LLMs for Serendipity Discovery in Knowledge Graphs:\\A Case for Drug Repurposing}
\author{
   Mengying Wang\equalcontrib, Chenhui Ma\equalcontrib, Ao Jiao\equalcontrib, Tuo Liang, Pengjun Lu, Shrinidhi Hegde,\\ Yu Yin, Evren Gurkan-Cavusoglu, Yinghui Wu
}
\affiliations{
    Case Western Reserve University, Cleveland, OH, USA\\

    \{mxw767, cxm590, axj770, txl859, pxl465, sxh1426, yxy1421, exg44, yxw1650\}@case.edu
}

\usepackage{bibentry}

\input{commands}

\begin{document}

\maketitle

\input{sections/abstract}
\input{sections/intro}
\input{sections/preliminary}

\input{sections/sscore}

\input{sections/benchmark}
\input{sections/pipeline}
\input{sections/experiment}

\input{sections/conclusion}

\section*{Acknowledgements}
This work is supported by NSF under OAC-2104007. 
We gratefully acknowledge the support of Dr. Rıza Mert Çetik and Dr. Sıla Çetik in the design and annotation of the QA dataset curated in this study.
We also acknowledge the HPC resources at CWRU for supporting large-scale graph processing and embedding computation.

\bibliography{papers}

\newpage
\input{sections/appendix}

\end{document}

%% file: commands.tex
\usepackage{graphicx}
\usepackage{amsmath,amsfonts}
\usepackage{textcomp}
\usepackage{latexsym}
\usepackage{color}
\usepackage{epsfig}
\usepackage{xspace}

\usepackage{pbox}
\usepackage{caption}
\usepackage{epsfig}
\usepackage{url}
\usepackage{mathrsfs}
\usepackage{booktabs}
\usepackage{adjustbox}
\usepackage{array}
\usepackage{multirow}
\usepackage{empheq}
\usepackage{tcolorbox}
\usepackage{enumitem}
\setlist{topsep=0pt,noitemsep} \setitemize[1]{label=$\circ$}
\usepackage{lipsum}
\usepackage{lineno}
\usepackage[utf8]{inputenc}

\newcommand{\eat}[1]{}
\newcommand{\stab}{\rule{0pt}{8pt}\\[-1.6ex]}

\newcommand{\sstab}{\rule{0pt}{8pt}\\[-2.4ex]}

\newcommand{\bi}{\begin{itemize}}
\newcommand{\ei}{\end{itemize}}
\newenvironment{tbi}{\begin{itemize}
        \setlength{\topsep}{1.5ex}\setlength{\itemsep}{0ex}}
        {\end{itemize}\vspace{-0.5ex}}

\newcommand{\be}{\begin{enumerate}}
\newcommand{\ee}{\end{enumerate}}
\newcommand{\beqn}{\begin{eqnarray*}}
\newcommand{\eeqn}{\end{eqnarray*}}

\newcommand{\stitle}[1]{\vspace{1.5ex}\noindent{\bf #1}}

\newcommand{\ie}{\emph{i.e.,}\xspace}
\newcommand{\eg}{\emph{e.g.,}\xspace}

\newcommand{\kw}[1]{{\ensuremath {\mathsf{#1}}}\xspace}

\newcounter{ccc}

\newcommand{\A}{{\mathcal A}}

\newcommand{\E}{{\mathcal E}}

\newcommand{\V}{{\mathcal V}}

\newcommand{\G}{{\mathcal G}}
\renewcommand{\L}{{\mathcal L}}

\newcommand{\eop}{\hspace*{\fill}\mbox{$\Box$}}     
\newcounter{example}
\renewcommand{\theexample}{\arabic{example}}
\newenvironment{example}{
        \vspace{1.5ex}
        \refstepcounter{example}
        {\noindent\bf Example \theexample:}}{
        \eop\vspace{1.5ex}}

\newcommand{\nthesection}{\arabic{section}}
\newcounter{definition}[section]

\newcounter{alg}[section]
\renewcommand{\thealg}{\nthesection.\arabic{alg}}

\newcounter{arule}
\renewcommand{\thearule}{\arabic{arule}}

\newcounter{claim}

\renewcommand{\texttt}[1]{{\small\textsf{#1}}}

\definecolor{gray}{rgb}{0.5,0.5,0.5}

\newcommand{\eetitle}[1]{\vspace{0.8ex}\noindent{\em\underline{#1}}}

\newcommand{\warn}[1]{{\color{red}{#1}}}

\DeclareMathOperator*{\argmax}{arg\,max}

\renewcommand{\partial}{\colorbox{yellow!30}{Partial}}

\newcommand{\serenqa}{\kw{SerenQA}}

\newcommand{\kg}{\kw{KG}}
\newcommand{\kgqa}{\kw{KGQA}}
\newcommand{\rns}{\kw{RNS}}

\newcommand{\llm}{\kw{LLM}}
\newcommand{\llms}{\kw{LLMs}}

\newcommand{\gpt}{\kw{GPT}-\kw{4o}}
\newcommand{\claude}{\kw{Claude}-\kw{3.5}-\kw{Haiku}}

\newcommand{\dsh}{\kw{DeepSeek}-\kw{V3}}
\newcommand{\lal}{\kw{Llama}-\kw{3.3}-\kw{70B}}
\newcommand{\dsl}{\kw{DeepSeek}-\kw{R1}-\kw{70B}}
\newcommand{\qwm}{\kw{Qwen3}-\kw{32B}}
\newcommand{\dsm}{\kw{DeepSeek}-\kw{R1}-\kw{32B}}
\newcommand{\qws}{\kw{Qwen3}-\kw{8B}}
\newcommand{\dss}{\kw{DeepSeek}-\kw{R1}-\kw{8B}}
\newcommand{\qwt}{\kw{Qwen3}-\kw{1.7B}}
\newcommand{\dst}{\kw{DeepSeek}-\kw{R1}-\kw{1.5B}}
\newcommand{\medl}{\kw{Med42}-\kw{V2}-\kw{70B}}
\newcommand{\meds}{\kw{Med42}-\kw{V2}-\kw{8B}}

\newcommand{\tdsl}{\kw{DeepSeek}-\kw{R1}-\kw{70B}}
\newcommand{\tlal}{\kw{Llama}-\kw{3.3}-\kw{70B}}
\newcommand{\tqwl}{\kw{Qwen}-\kw{2.5}-\kw{72B}}
\newcommand{\tmxm}{\kw{Mixtral}-\kw{8x7B}}
\newcommand{\tqwm}{\kw{Qwen}-\kw{2.5}-\kw{32B}}
\newcommand{\tgmm}{\kw{Gamma}-\kw{2}-\kw{27B}}
\newcommand{\tmss}{\kw{Mistral}-\kw{24B}}
\newcommand{\tqws}{\kw{Qwen}-\kw{2.5}-\kw{7B}}

\NewEnviron{charcount}{
    \def\charcount@tmp{\BODY}
    \StrSubstitute{\charcount@tmp}{ }{}[\charcount@tmp]
    \StrSubstitute{\charcount@tmp}{~}{}[\charcount@tmp]
    \StrLen{\charcount@tmp}[\templen]
    \BODY
    \par \textcolor{blue}{\\\textbf{Character Count (without spaces, newline counts as 1): \templen}}%
}

%% file: sections/abstract.tex
\begin{abstract}

Large Language Models (LLMs) have greatly advanced knowledge graph question answering (KGQA), yet existing systems are typically optimized for returning highly relevant but predictable answers.
A missing yet desired capacity is to exploit LLMs to suggest surprise and novel  (``serendipitious'') answers.
In this paper, we formally define the serendipity-aware KGQA task and propose the SerenQA framework to evaluate LLMs' ability to uncover unexpected insights in scientific KGQA tasks. 
SerenQA includes a rigorous serendipity metric based on relevance, novelty, and surprise, along with an expert-annotated benchmark derived from the Clinical Knowledge Graph, focused on drug repurposing. 
Additionally, it features a structured evaluation pipeline encompassing three subtasks: knowledge retrieval, subgraph reasoning, and serendipity exploration.
Our experiments reveal that while state-of-the-art LLMs perform well on retrieval, they still struggle to identify genuinely surprising and valuable discoveries, underscoring a significant room for future improvements. Our curated resources and extended version are released at: https://cwru-db-group.github.io/serenQA.

\end{abstract}

%% file: sections/intro.tex
\section{Introduction}
\label{sec:intro}

Large language models (LLMs) are rapidly advancing the bridge between natural language understanding and effective question answering.
\eat{
Recent AI 
techniques such as Retrieval-Augmented Generation 
(RAG), which grounds the output of LLMs by 
allowing them to access 
and incorporate information from external  
domain knowledge bases (such as 
knowledge graphs), further enhance them 
to provide domain-specific answers~\cite{}. 
Despite the success, existing LLMs and RAG systems 
are still providing
high relevance answers that the queryers 
may already be familiar with, and remain limited 
in finding new, unexpected answers that 
may, in turn, be more inspiring.} 
Significant efforts, such as domain-specific fine-tuning, prompt engineering, and Retrieval-Augmented Generation (RAG), have enabled LLMs to leverage external knowledge bases
to produce highly relevant and precise answers tailored to specialized research questions~\cite{le2024graphlingo}.
However, these systems often focus on returning information already familiar to experts, missing the crucial scientific capacity to uncover surprising connections that inspire new research directions~\cite{song2023advancements}.

``\textit{Serendipity}'', the art of luck and beneficial discovery, arises from both unexpected findings and the skill to recognize novel applications of such discoveries in various domains, serving as a catalyst for genuine scientific breakthroughs.
While serendipity has been studied in web search~\cite{huang2018learning} and recommender systems~\cite{tokutake2024can}, it remains largely unexplored in scientific question answering. Empowering LLMs with the ability to discover new knowledge from existing, valuable knowledge bases is thus a critical step towards true LLM-empowered scientific discovery.

\begin{figure}[tb!]
\centerline{\includegraphics[width =0.44\textwidth]{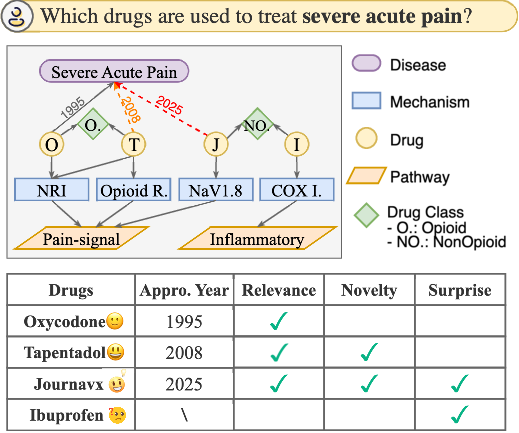}}
\centering
\caption{Suggesting Drugs that treat Severe Acute Pain: A Serendipitous case of Journavx.}
\label{fig:motivation}
\end{figure}

\begin{example}
\label{exa-motivation}
Fig.~\ref{fig:motivation} illustrates a \kgqa task to find drugs that can treat severe acute pain. 
There are four possible answers. (1) Opioids \eg \textbf{Oxycodone}, a well-known drug with recognized mechanism on targeting the $\mu$-opioid receptor within the pain-signaling pathway. 
(2) \textbf{Tapentadol} (2008) expanded this paradigm by adding a dual mechanism, hence with increased novelty for the question. 
(3) \textbf{Journavx}, a first non-opioid analgesic for severe acute pain~\cite{journavx} approved by FDA in 2025. Journavx acts through a novel mechanism, selectively inhibiting NaV1.8 sodium channels in peripheral pain-sensing neurons. Surprisingly, with this paradigm shift and different targets, it remains relevant by sharing the broader pain-signal pathway context with opioids. Hence 
it is a ``serendipitous'' result in the KGQA search, in terms of 
relevance, novelty, and an unexpected answer, which may inspire new medical research directions. 
(4) \textbf{Ibuprofen}, in contrast, works through the classical inflammatory COX inhibition pathway, targeting mild-to-moderate pain and thus showing low embedding relevance and novelty, while suggesting Ibuprofen for severe acute pain would still be surprising.
\end{example}

\eat{
In this context, recent RAG 
techniques such as graphRAG~\cite{peng2024graph} 
allows LLMs to consult an external 
domain knowledge graph (\kg) to retrieve  
answers that are more accurate. Nevertheless, enforcing strong relevance from \kg 
alone could be an overkill for serendipitous, 
interesting answers, as remarked in the 
example. 
}

``{\em Can LLMs, while enhanced by 
domain KGs, suggest serendipitous answers 
for domain sciences?''} 
This paper makes a first step to investigate the potential of 
LLMs to surface serendipitous discoveries within scientific \kgqa, with a focus on drug repurposing, which is a cornerstone of medical research.
We address three core research questions: 
\tbi 
\item  (\textbf{RQ1}): How may ``serendipity'' be characterized and quantitatively measured for scientific \kgqa tasks? 
\item  (\textbf{RQ2}): What roles could LLMs play 
for serendipity discovery in domain science \kgqa? 
\item  (\textbf{RQ3}): How to evaluate state-of-the-art LLMs, 
and what are their performances in serendipity discovery?
\ei 


To this end, we introduce the \serenqa framework 
designed to systematically evaluate the ability of LLMs to uncover serendipitous answers within the context of \kgqa. It includes three core components (shown in Fig.~\ref{fig:framework}):


\tbi
    \item \textbf{Serendipity Metric (\rns)}: A rigorous, graph-based measure capturing \textbf{R}elevance, \textbf{N}ovelty, and \textbf{S}urpriseness in \kgqa answers, justified by an \textit{axiomatic} analysis that clarifies the trade-offs and properties.  
    \item \textbf{Serendipity-aware Benchmark}: An expert-annotated KGQA dataset for drug repurposing, based on the Clinic Knowledge Graph~\cite{Santos2022May}. It features curated question-answer pairs and explicit serendipity annotations for fine-grained evaluation.
    \item \textbf{Assessment Pipeline}: 
     A principled and reproducible three-phase workflow that systematically evaluates LLMs’ roles in serendipitous discovery. It decomposes the task into knowledge retrieval, reasoning, and exploratory search, providing insights into model capabilities and limitations in scientific knowledge discovery.
\ei

\vspace{.5ex}
We performed extensive experiments with various LLMs across different scales, demonstrating that while frontier models excel in knowledge retrieval tasks, nearly all models struggle significantly in serendipity exploration, highlighting inherent challenges and opportunities in this area.

\stitle{Related works}. We categorize related works as follows. 

\eetitle{Serendipity-Driven Knowledge Exploration}
Serendipity, defined as an unexpected yet valuable discovery, has emerged as a crucial goal in recommender systems and knowledge exploration~\cite{bordino2013penguins}. 
Recent studies have leveraged LLMs to generate and evaluate serendipitous recommendations through advanced prompt engineering~\cite{fu2024art} or by aligning model outputs with human preferences~\cite{xi2025bursting}. 
Notably, existing approaches primarily rely on subjective human annotation, LLM self-evaluation, or comparisons against benchmark groundtruths for serendipity evaluation.
In contrast,  we propose a graph-based serendipity measure (\rns), which transforms the knowledge graph (KG) into a probability matrix~\cite{dehmer2011history}, enabling an information-theoretic quantification of various subjective aspects of serendipity, resulting in a more rigorous evaluation.

\vspace{.5ex}
\eetitle{LLM-Augmented Novelty Detection}. 
LLMs are increasingly seen as creative partners that can accelerate scientific discovery across disciplines~\cite{ai4science2023impact}.
By mining vast knowledge and generating hypotheses, LLMs can propose novel research ideas or unexpected connections that human experts might overlook~\cite{Si2025Can}.
Despite these efforts, the community still lacks a more comprehensive understanding and benchmark datasets specifically designed to assess serendipitous discoveries.
To address this gap, we present a drug repurposing \kgqa dataset
which enables a systematic and objective assessment of serendipitous knowledge exploration.

\serenqa is the first reproducible and extensible 
framework for advancing serendipity discovery 
in drug repurposing. 
We advocate its broader application to facilitate new research opportunities in scientific \kgqa tasks.

%% file: sections/preliminary.tex
\section{Serendipitous Assessment with KGQA}
\label{sec-pre}

Below, we define relevant concepts and core notations: 

\begin{figure*}[tb!]
    \centering
    \includegraphics[width = 0.85\textwidth]{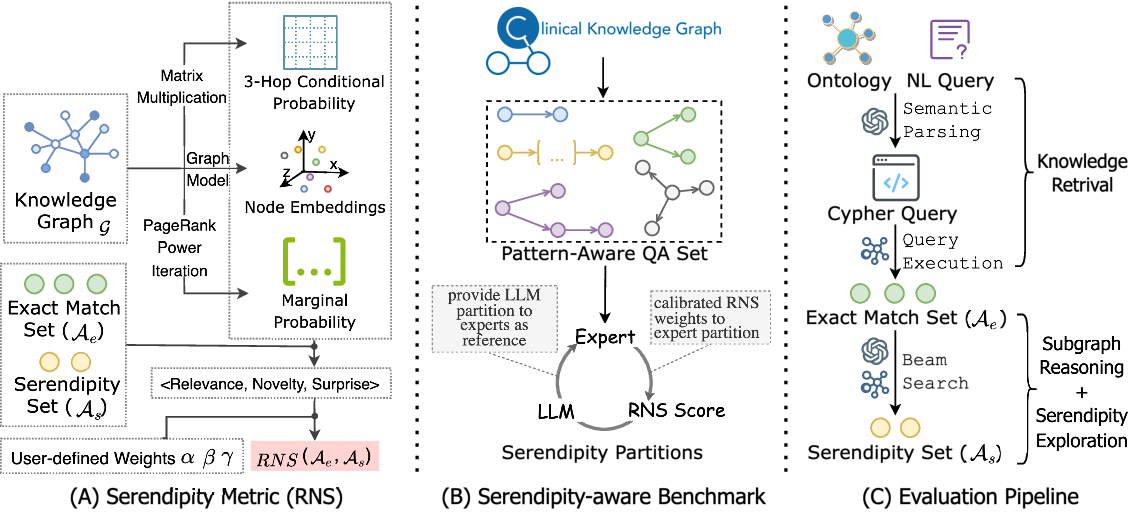}
    \centering
    \caption{\small{\serenqa Framework. 
    \textbf{(A)}: Computing \rns score for partition $(\A_e, \A_s)$ form $\G$; (Sec.~\ref{sec:quantifiy}). \textbf{(B)}: Constructing \serenqa dataset from ClinicalKG; 
(Sec.~\ref{sec:benchmark}). 
    \textbf{(C)}: For an NL query, our pipeline retrieves $\A_e$ from $\G$ and explores $\A_s$ from $\A_e$ with beam search. 
    \vspace{-2ex}
    (Sec.~\ref{sec:pipeline}).}}
    \label{fig:framework}
\end{figure*}

\subsection{Serendipity-aware \kgqa}

\serenqa performs \llm assessment by processing a pipeline 
of {\em serendipity-aware \kgqa}.  
Given a natural language (NL) question $Q$, a large language model $\L$, a directed, multigraph $\G=(\V,\E)$, where $\V$ is the set of entities with size $V = |\V|$, and $\E$ is the set of relations with size $E = |\E|$, a serendipity-aware \kgqa system 
returns an answer set as an ordered partition $\A=(\A_e, \A_s)$, where:
\begin{itemize}
\item $\A_e$: the \textbf{existing answer set}, containing answers explicitly supported by facts in $\G$;
\item $\A_s$: the \textbf{serendipity answer set}, containing answers that are relevant but extend beyond direct explicit knowledge, revealing novel and unexpected relationships in $\G$.
\end{itemize}
such that $\A_e\cup\A_s\subseteq\V$ and $\A_e \cap \A_s = \emptyset$. We define $|\A| = |\A_e \cup \A_s|$ as the total size of the answer set.

This serendipity-aware setup is motivated by the real-world scientific discovery process, which frequently involves uncovering not only established knowledge ($\A_e$) but also insightful and surprising associations ($\A_s$), potentially leading to innovative research opportunities, such as novel drug repurposing.
Knowledge graphs are particularly suitable for this task due to their structured representation of interconnected entities and relations, enabling systematic exploration of indirect and surprising relationships.

\subsection{Graph-specified Serendipity Formulation}
\label{sec:pre:score}

To rigorously quantify serendipity, we define a graph-based serendipity measure (\rns), which quantifies how effectively a serendipity answer set $\A_s$ for a given question $Q$ provides relevant yet novel and surprising insights beyond the explicit answer set $\A_e$.
Intuitively, serendipity is a composite experience, encompassing multiple dimensions simultaneously~\cite{niu2017framework}. 
Formally, we define the \rns score as a weighted combination of three 
perspectives: relevance, novelty, and surprise, which can be flexibly adjusted to suit user preferences. Given an answer set $\A=(\A_e, \A_s)$, the serendipity score is computed as:
$$\rns(\A_e,\A_s) = \alpha\,R(\A_e,\A_s)+\beta\,N(\A_e,\A_s)+\gamma\,S(\A_e,\A_s)
$$

\sstab
- $R$ (\textbf{Relative Relevance}): context similarity of $\A_e$ and $\A_s$;

\sstab
- $N$ (\textbf{Relative Novelty}): new information in $\A_s$ beyond $\A_e$;

\sstab
- $S$ (\textbf{Relative Surprise}): unpredictability of $\A_s$ given $\A_e$.

The weights $\alpha, \beta, \gamma$ can be tuned to user preference; recommended defaults are fit to expert evaluations.
Details of the metric and its computation are described in Sec~\ref{sec:quantifiy}.


In the following sections, we detail how the \serenqa framework establishes a unified benchmark, dataset, and evaluation protocols specifically designed to assess LLM capabilities in serendipitous knowledge discovery tasks, particularly in the critical area of drug repurposing.

%% file: sections/sscore.tex
\vspace{-1ex}
\section{Serendipity Quantification} 
\label{sec:quantifiy}

Quantifying serendipity is inherently challenging due to its abstract and subjective nature.
As discussed in Sec~\ref{sec:intro}, existing methods often rely heavily on subjective human annotations or LLM-generated evaluations, which suffer from limitations like poor interpretability, scalability issues, and potential biases.
To overcome these, we introduce an information-theoretic approach enabling \textit{scalable}, \textit{interpretable}, and \textit{reproducible} serendipity evaluations.

\vspace{-1ex}
\subsection{Serendipity: A characterization}

To align with human intuition about ``Serendipity'' while allowing for rigorous quantification, as introduced in Sec~\ref{sec:pre:score}, we specifically decompose it into three complementary dimensions: \textit{Relevance}, \textit{Novelty}, and \textit{Surprise}. 
For an answer set $\A = (\A_e, \A_s)$ to a query $Q$, we define the \textbf{Serendipity Score (\rns)} as a weighted combination of the relative measures between $\A_s$ and $\A_e$, with user-configurable weights to accommodate different preferences or application scenarios. 
Each dimension is adapted to well-established information-theoretic measures, as described below:

\stitle{Relative Relevance}. 
We compute relative Relevance ($R$) as the average normalized Euclidean distance ($d(\cdot)$) between the 
GCN embeddings of entities in $\A_s$ and $\A_e$: 
\begin{small}
$$
R(\A_e, \A_s) = - \frac{\sum_{i \in \A_s, j\in \A_e} d(n_i, n_j)}{|\A_s| |\A_e|} 
$$
\end{small}
where $n_i$ (resp. $n_j$) refers to the embedding 
of the entity $i\in A_s$ (resp. $j\in \A_e$). 
A larger distance reflects greater contextual difference, indicating $\A_s$ belongs to more distinct clusters in $\G$ and may diverge from the core context of $Q$.

\stitle{Relative Novelty}. Relative Novelty ($N$) is derived from a mutual-information-based score between the existing and serendipity sets. For a partition $(\A_e, \A_s)$,
we define $N(\A_e, \A_s)$ = 1- $MI(\A_e, \A_s)$,
where $MI(\A_s, \A_e)$ measures the shared amount of information between $\A_s$ and $\A_e$, and is given by: 
\begin{small}
$$MI(\A_e, \A_s) = \sum_{i \in \A_e} P(i) \sum_{j \in \A_s} P(j | i) \log \frac{P(j | i)}{P(j)}$$
\end{small}
A higher N score indicates 
$\A_s$ are less redundant given $\A_e$. 
\eat{more unexpected information}

\stitle{Relative Surpriseness}. Relative Surprise ($S$) is quantified via Jensen–Shannon divergence (JSD) between entity distributions $P_s$ and $P_e$, which are the accumulated probability distributions over entities in $\A_s$ and $\A_e$, respectively:
\begin{small}
$$S(\A_e, \A_s) =\frac{1}{2} (D_{KL}(P_s \| P_{Mix}) + D_{KL}(P_e \| P_{Mix})) $$
\end{small}
where $D_{KL}(\cdot\|\cdot)$ is the Kullback–Leibler divergence~\cite{kullback1951kullback}, 
and $P_{Mix} = \frac{1}{2}(P_s + P_e)$.

Given $\A_e$, a \textit{higher} $\rns$ indicates a ``more'' serendipitous set $\A_s$ 
with 
greater diverse, novel 
and surprise entities that cannot be inferred from $\A_e$, as 
exemplified by ``Journavx'', the first non-opioid analgesic for severe acute pain (Exp.~\ref{exa-motivation}). 



\subsection{Cost-effective Graph Probabilistic Modeling} 
\label{sec:modeling}

Cost-effective graph probabilistic models ($P(\cdot)$) 
is crucial for efficient \rns computation.
We present the detailed models, justified by an axiomization analysis.

\stitle{3-Hop Conditional Probability}.
Serendipitous findings may from indirect, multi-hop connections. Thus, we consider a multi-hop conditional probability matrix $M$ that aggregates transition probabilities across both direct and indirect relations to capture a global probabilistic propagation. 
Empirically, $99\%$ of serendipity answers in our datasets are reachable from existing answers within three hops, prompting our 
analysis to up to $3$-hop neighbors of entities in $\G$. 


Given graph $\G$, we initialize $M$ as a weighted 
matrix $M$, with $M_{ij}$ the number of links from node $i$ to $j$. 
We normalize $M$ to obtain the one-hop transition probabilities that ensures row-stochasticity: $P_1(j|i)$ = $\frac{M_{ij}}{\sum_{k \in \E} M_{ik}}$. 
The $k$-hop conditional probability matrix $P_k$ is 
\begin{small}
computed as:
$$
P_k = \sum_{h=1}^{k} \alpha_h P_1^h, 
\quad  \alpha_h = \frac{h}{\sum_{h=1}^{k} h} 
$$ 
\end{small}
where $P_1^h$ represents the probability of reaching a node in  $h$  hops, and weights $\alpha_h$ increases for larger $h$ to prioritize longer connections. 
We can justify that $P_k$ 
consistently satisfies the necessary constraints of a transition matrix:
\tbi
\item \textit{Non-negativity}: $(P_k)_{ij} \geq 0$ for all $(i, j)$,
\item \textit{Row-Stochastic Property}: $\sum_{j} (P_k)_{ij} = 1$  for all $i$.
\ei

\eetitle{Cost Analysis.} 
Constructing $M$ takes $\mathcal{O}(V^2)$ for dense graphs. 
Traditional $P_3$ computation\footnote{While we make a case for $3$-hop queries here, our discussion readily extends to $k$-hop queries for $k\geq 3$.} via graph traversal is $\mathcal{O}(V^4)$. 
We employ Divide-and-Conquer optimized matrix multiplication~\cite{strassen1969gaussian} and parallel computation with $t$ processors, reducing the cost to $\mathcal{O}(V^{\log_2^7}/t)$.

\stitle{Marginal Probability}. 
The marginal probability $\mathbf{P}(i)$ quantifies steady-state node probabilities at node $i$ under the law of total probability: $\mathbf{P}(i) = \sum_j P_3(i | j) \mathbf{P}(j)$.  
This leads to the linear system representation:
\begin{small}
$$(I - P_3^T) \mathbf{P} = 0, \quad \sum_i \mathbf{P}(i) = 1$$
\end{small}
which can be solved by matrix inversion in $\mathcal{O}(V^3)$ time. To further reduce the cost, we approximate 
the computation with a PageRank-style damped iteration:
\begin{small}
$$\mathbf{P}_{t+1} = \lambda P_3^T \mathbf{P}_t + (1 - \lambda) \mathbf{P}_0
\vspace{-.5ex}
$$
\end{small}
where $\mathbf{P}_0$ is an initial probability distribution, set uniformly as $\frac{1}{V}$, ensuring convergence even on disconnected graphs. This reduces the cost in $\mathcal{O}(V^2\log V)$ time.

We remark that the probabilistic matrices are computed 
``once for all'' and are shared for multiple queries, 
and readily adapt to different domain graphs. 

Further analyses are included in the 
Appendix~\ref{sec:app:score}.

\stitle{Axiomization Analysis}. We further justify that $\rns$ is a proper serendipity measure for \kgqa tasks through the following axiomatic analysis. For any query and a corresponding 
retrieved, fixed existing set $\A_e$, consider 
an optimization process that finds an 
optimal serendipitous set $\A^*_s$ with at most $K$ entities, \ie $A_s^*$ = $\argmax_{|A_s|\leq K}$ $\rns(\A_e,\A_s)$. 
We can show that $\rns$ satisfies the following properties:
\tbi 
\item (\textbf{Scale invariance}). $\A^*_s$ remains to maximize $\rns$ 
even if $R$, $N$ or $S$ are scaled by a constant. This ensures the invarance of $\A^*_s$ under $\rns$ 
measure regardless of 
how the user preference ($\alpha$, $\beta$, $\gamma$) changes. 
\item (\textbf{Consistency}). Making the $R$, $N$, $S$ larger 
(resp. smaller) for any entities in $\A_e$ (resp. $\A_s$) 
does not change the ranking of entities 
in $\A^*_s$ in terms of \rns.  
\item (\textbf{Non-monotonicity}). 
$\rns(\A_e, \A_s)\not\leq \rns(\A_e, \A'_s)$ 
if $|\A_s|\leq |\A'_s|$. 
Indeed, larger answer sets do \textit{not} necessarily 
indicate that they are more ``serendipitous'' in practice. 
\item (\textbf{Independence}). $\rns$ is only determined by 
the embeddings of entities from $\A_s\cup \A_e$. 
No information from entities not seen in $\A$ 
can affect the serendipitous of $\A_e$. This 
justifies $\rns$ for serendipity in a 
pragmatic ``semi-closed world'' 
assumption, striking a balance between 
a challenging open-world analysis ($\A_s$ can be 
infinite) and a
rigorous, overkilling closed world 
($\A_s$ = $\emptyset$)
setting. 
\ei

%% file: sections/benchmark.tex
\section{Serendipity-aware Benchmark}
\label{sec:benchmark}

The proposed \rns measure enables quantitative assessment of serendipity within any answer set $(\A_e, \A_s)$ derived from a graph $\G$. 
Yet scoring alone is insufficient: assessing cornerstone steps such as retrieving and reasoning demands a benchmark with authoritative groundtruth serendipity
answer set. 
We therefore introduce a drug-repurposing benchmark that supports both standard KGQA tasks and serendipity-aware evaluations, 
giving the fine-grained supervision required for end-to-end assessment.

\eat{
Serendipity evaluation, including 
cornerstone steps such as retrieving, reasoning 
and quality verification, requires a robust benchmark that best exploits available knowledge sources. 
We next introduce a comprehensive serendipity evaluation benchmark in the context of drug repurposing. The benchmark supports both standard \kgqa tasks and serendipity 
assessment inquiries. 
}

\subsection{QA Set Construction} 
\label{sec:base}

Our benchmark is built upon the Clinical Knowledge Graph (CKG)~\cite{Santos2022May}, a widely recognized biomedical resource containing extensive data on drug, gene, and disease interactions. 
Our focus is on drug repurposing, which is a critical research task aimed at identifying novel therapeutic uses of existing drugs~\cite{pushpakom2019drug}.

Our dataset supports typical \kgqa tasks through a contextualized query scenario that consists of standardized configuration including \textit{expert-verified}, scientifically meaningful NL queries, 
their structured graph (Cypher) counterparts 
with query components that are explicitly 
annotated with their semantics, and grounded 
and validated answer sets. 
Unlike its peer NL-only benchmark datasets 
in KGQA, it 
couples each NL query to a distinct, validated 
``ground truth'', structured graph query, thereby 
reducing ambiguity and mitigating possible 
semantic redundancy. 
It also explicitly annotates graph patterns, such as multi-hop and intersection queries, to reflect realistic query complexities in scientific inquiry. 
Dataset statistics are summarized in Table~\ref{tab:dataset-stats}. We 
present details of graph queries in 
Appendix~\ref{sec:app:dataset}.


\begin{table}[tb!]
\centering
\begin{small}
\begin{tabular}{lr}
\toprule
\textbf{Statistic} & \textbf{Value} \\
\midrule
Number of Distinct Queries & 1529 \\
Number of Relations in $\G$ ($E$) & 201,704,256 \\
Number of Entities in $\G$ ($V$) & 15,430,157 \\
Number of Graph Pattern Types & $9$ \\
Avg. Answer Set Size ($|\mathcal{A}|$ per query) & 4.04\\
Number of Experts for NL Query Verification & $4$ \\
Number of Experts for Serendipity Annotation & $6$ \\
\bottomrule
\end{tabular}
\end{small}
\caption{Dataset Statistics of \serenqa Benchmark.}
\vspace{-3ex}
\label{tab:dataset-stats}
\end{table}

\subsection{Answer Set Construction}
\label{sec:partition}

To reliably establish ground-truth serendipity sets, we start with the latest version of Clinic KG, denoted as $\G_c$.
For each query $Q$, we initially obtain its complete candidate answer set $\A_c$ from $\G_c$. 
We then partition it into an existing set $\A_e$ 
and a serendipity set $\A_s$ 
, with $\A_e \cap \A_s = \emptyset$ and $\A_e \cup \A_s = \A_c$.
We apply three distinct partitioning strategies:

\stitle{LLM Ensemble}. 
Following established practices, 
we prompt four state-of-the-art LLMs to assign a “serendipity score” to each candidate answer. 
For every query, entities in the complete answer set $\A_c$ are ranked by their average LLM score; 
the top $20\%$ are collected as the serendipity set $\A_s$, 
and the reminder form $\A_e$.


\vspace{-1ex}

\stitle{Expert Crowdsourced}.
We engaged a team of $6$ domain experts (three physicians, one pharmaceutical scientist, and two trained medical model annotators) via an online questionnaire
~\cite{questionnaire}. 
They were requested to refine the rankings from \llms.
The questionnaire is accepting continuous responses from human experts. 

\stitle{RNS Guided}. 
With the justified \rns metric (Sec.\ref{sec:quantifiy}) we treat serendipity partitioning as:
$$ 
\max_{\A_e, \A_s} \rns(\A_e, \A_s), \quad \text{s.t. } |\A_s| = b, b = \max(1, \lfloor 0.2|\A_c| \rfloor)
$$
Starting from an initial partition, we apply the greedy-swap algorithm in Algorithm~\ref {alg:greedy-swap} to (approximately) 
compute an optimal answer set $\A_s$ in $\A_c$.
The algorithm iteratively swaps entity pairs between $\A_e$ and $\A_s$ that yield the greatest improvement in a marginal gain of \rns, continuing until no further improvement is possible. 
Each iteration has a complexity of $\mathcal{O}(|\A|^2)$. We found in our tests that $\A_e$ 
usually contains a few entities (on average 4; see Table~\ref{tab:dataset-stats}), 
And the algorithm is quite fast in practice. 
During that, we calibrated the \rns weights to align with the expert-crowdsourced partitions for consistency and fair assessment. 

\begin{algorithm}[tb!]
\caption{Greedy Swap for \rns–Guided Optimization}
\label{alg:greedy-swap}
\textbf{Input}: initial partition $(\A_e^0,\A_s^0)$;\\  
pre-computed probability matrices $P_3$, $\mathbf{P}$ for graph $\G$ \\
\textbf{Output}: optimized partition $(\A_e,\A_s)$ \\
\begin{algorithmic}[1]
\State \textbf{set} $(\A_e,\A_s):=(\A_e^0,\A_s^0)$,\; 
$\tau = \rns(\A_e,\A_s)$
\While{\textbf{true}}
    \State \textbf{set} $\Delta_{\max}:=0;\;(i^\ast,j^\ast):=\texttt{null}$
    \For{$i\in\A_s$}
        \For{$j\in\A_e$}
            \State $\A_s' := (\A_s\!\setminus\!\{i\})\cup\{j\}$,\;
            $\A_e' := (\A_e\!\setminus\!\{j\})\cup\{i\}$
            \State $\Delta:=\rns(\A_e',\A_s')-\tau$
            \If{$\Delta>\Delta_{\max}$}
                \State $\Delta_{\max}:=\Delta$;\;$(i^\ast,j^\ast):=(i,j)$
            \EndIf
        \EndFor
    \EndFor
    \If{$\Delta_{\max}=0$}\; \textbf{break}; \EndIf
    \State $\A_s := (\A_s\!\setminus\!\{i^\ast\})\cup\{j^\ast\}$,\;
            $\A_e := (\A_e\!\setminus\!\{j^\ast\})\cup\{i^\ast\}$
    \State $\tau := \tau + \Delta_{\max}$
\EndWhile
\State \textbf{return} $(\A_e,\A_s)$
\end{algorithmic}
\end{algorithm}



For each partitioning result, we construct $\G$ by removing selected edges from $\G_c$, ensuring that for each query $Q$, entities in $\A_e$ remain derivable from $\G$, while entities in $\A_s$ become inaccessible. This creates a controlled evaluation environment aligned with problem definitions (Sec.~\ref{sec-pre}).

%% file: sections/pipeline.tex
\vspace{-1ex}
\section{Evaluation Pipeline} 
\label{sec:pipeline}

We next introduce our evaluation pipeline (Fig.~\ref{fig:framework}(C)), which systematically evaluates the serendipity discovery capabilities of LLMs using our curated serendipity-aware benchmark. 
The pipeline is modularized into three highly correlated tasks, each of which independently measures a specific, ``cornerstone'' aspect of an LLM's role and performance on serendipity discovery in scientific \kgqa tasks. 


\vspace{-1ex}

\stitle{Knowledge Retrieval}. 
In this task, LLM translates an NL question $Q$ into a Cypher query $C$ to retrieve an answer set $\A_e$ from the knowledge graph $\G$. 
The performances are evaluated by comparing the accuracies 
of the retrieved answer set $\A_e$ against the ground truth. 
Additionally, the performances across different query patterns (such as one-hop, two-hop, and intersection queries) are compared to evaluate the LLM's capability to handle varying levels of query complexity and structural diversity.

\vspace{-1ex}

\stitle{Subgraph Reasoning}.
This task evaluates the LLM's capability to interpret and concisely summarize the retrieved answer of 
a graph-structured query $C$ in a 
knowledge graph (as a subgraph) into domain-aware natural language answers. 
The generated summaries provide essential contextual support for subsequent serendipity exploration tasks, 
requiring nuanced biomedical understanding and logical reasoning. 

\vspace{-1ex}

\stitle{Serendipity Exploration}.
This third (final) task evaluates the LLMs' proactive ability to uncover serendipity entities $\A_s$ through an LLM-guided beam search from $\A_e$. 
Given a beam width $w$, we prompt LLM to select the top-$w$ nodes at each step from the candidate list as the next target nodes based on criteria such as supporting evidence, interaction strength, biological effect direction, and their expression level. The model further determines whether to continue exploration based on relevance and potential novelty. 
This task assesses the LLM's ability to use biomedical knowledge and contextual search to effectively navigate serendipitous discovery while balancing depth and breadth in exploration.
We remark that the serendipity set $\A_s$ produced in this section is the pipeline’s output at evaluation time;
in contrast, the $\A_s$ defined in Sec.~\ref{sec:benchmark} is the benchmark ground-truth constructed for scoring.
More details are provided in 
Appendix~\ref{sec:app:ppl}.

%% file: sections/experiment.tex
\section{Experiments}
\label{sec:exp}

\subsection{Experiment Setting}
\label{sec:exp: setting}

\begin{table*}[tb!]
  \centering
  \small
  \begin{adjustbox}{max width=\textwidth}
    \centering
    \begin{tabular}{ccccccccccccc}
      \toprule
      Model & \multicolumn{3}{c}{\textbf{One-Hop}} & \multicolumn{3}{c}{\textbf{Two-Hop}} & \multicolumn{3}{c}{\textbf{Multiple(3+)-Hop}} & \multicolumn{3}{c}{\textbf{Intersection}}\\
      \cmidrule(lr){2-4} \cmidrule(lr){5-7} \cmidrule(lr){8-10} \cmidrule(lr){11-13} 
      & Hit($\%$) & F1($\%$) & Exe.($\%$)
      & Hit($\%$) & F1($\%$) & Exe.($\%$)
      & Hit($\%$) & F1($\%$) & Exe.($\%$)
      & Hit($\%$) & F1($\%$) & Exe.($\%$)\\
      \midrule
      \dsh &\textbf{20.45} &\textbf{78.71} & \underline{72.88} & 3.46 & 10.71 & 9.86 & 1.97 & 6.22 & 6.55 & 2.64 & \underline{7.15} & 8.03 \\
      \gpt & 19.71 & \underline{77.16} & 60.17 & 2.08 & 6.36 & 7.89 & 1.40 & 4.20 & 4.85 & 1.56 & 4.65 & 5.21\\
      \claude & 13.28 & 48.54 & 48.73 & 9.78 & \underline{39.01} & 32.89 & \textbf{4.43} & \underline{8.64} & \textbf{14.08} & 1.38 & 3.90 & 4.66 \\
      \midrule
      \lal & 19.28 & 70.67 & \underline{74.58} & \textbf{16.63} & \textbf{44.34} & \textbf{56.57} & \underline{2.98} & \textbf{10.16} & 11.89 & \textbf{4.80} & \textbf{9.60} & \underline{16.05} \\
      \dsl & \underline{19.87} & 69.07 & \textbf{80.08} & \underline{12.03} & 37.00 & \underline{43.42} & 2.97 & 8.06 & \underline{13.11} & \underline{3.49} & 6.16 & \textbf{16.46} \\
      \medl & 18.34 & 69.43 &  69.92 & 5.92 & 19.12 & 19.74 & 0.23 & 0.51 & 1.21 & 0.08 & 0.13 & 0.68\\
      \midrule
      \qwm & 0.37 & 1.27 & 1.27& 0.16 & 0.65 & 0.65 & 0.24 & 0.36 & 0.48 & 0.00 & 0.00 & 0.00 \\
      \dsm &17.90 &65.23 &68.22&3.06&5.72&7.24&1.87&4.50&5.58&0.79&1.84&3.16 \\
      \midrule
      \qws &10.07 &37.24 &39.83&0.98&2.87&3.95&0.90&2.01&4.85&1.58&1.91&5.62 \\
      \dss &1.27 &3.41 &5.51&0.00&0.00&0.00&0.04&0.24&0.24&0.00&0.00&0.00 \\
      \meds &8.11 &23.90 &49.15 &1.05&3.31 & 3.97&1.71& 4.07& 4.12 & 0.04&0.13&0.14 \\
      \midrule
      \qwt &0.84 &3.72 &11.86&0.65&1.98&3.29&0.00&0.00&0.24&1.08&1.56&2.74 \\
      \dst &0.00 &0.00 &0.00 &0.00 &0.00 &0.00 &0.00 &0.00 &0.00 &0.00 &0.00 &0.00 \\
      \bottomrule
    \end{tabular}
    \end{adjustbox}
    \caption{Knowledge Retrieval ($T1$), Best scores are \textbf{bolded}, second best are \underline{underlined}}
    \vspace{-1ex}
    \label{table:task1}
\end{table*}

We conduct experiments using the benchmark introduced in Sec.~\ref{sec:benchmark}, 
and evaluated LLMs across multiple scales, from frontier models with billions of parameters to smaller variants ($~1B$ parameters).
Experimental results are presented in Tables~\ref{table:task1}–\ref{table:task3},  including three evaluation tasks within our pipeline: $T1$ (Knowledge Retrieval), $T2$ (Subgraph Reasoning) and $T_3$ (Serendipity Exploration). 

\stitle{Evaluation metrics}. Table~\ref{table:task1} ($T1$) reports \textit{F1 scores}, \textit{Executability} (percentage of error-free queries), and \textit{Hit Rate}($|\A_e \cap \A'_e|/|\A_e|$), categorized by query patterns; and Table~\ref{table:task3} ($T2$, $T3$)  reports their performances on three ground-truth partitions (LLM-Ensemble, Expert-Crowdsourced, \rns-Guided). 
During beam search (beam width $30$, maximum depth $3$), we employ one-shot learning by providing a single query with detailed ground-truth serendipity paths in the prompt, helping models understand exploration paths. In addition, $T2$ and $T3$ are measured with (a) {Subgraph Reasoning}:\textit{Faithful.} (1–5, LLM-judged, factual accuracy of summaries); \textit{Compre.} (1–5, LLM-judged, coverage of key graph elements); \textit{SerenCov} (0–1, fraction of serendipity paths explicitly mentioned). (b) {Serendipity Exploration}: 
\textit{Relevance} (1–5, LLM-judged alignment with groundtruth entities); \textit{TypeMatch} (0–1, the fraction of predicted entity types that match the ground truth types); and \textit{SerenHit} (0–1, match rate with groundtruth serendipity set). 

\stitle{Experiment Environment} We depoly our system on 5 x AWS c6a.24xlarge on-demand instances for distributed computation and 5 x c6a.xlarge instances as relation storage nodes, each node runs Ubuntu 22.04 with Docker and Redis 7.2, using mounted dump.rdb as readonly data source. The system supports 500 concurrent LLM reasoning tasks across distributed nodes via asyncio.






\begin{small}
\begin{table*}[tb!]
\centering
\begin{adjustbox}{max width=\textwidth}
\begin{tabular}{lccccccccc}
\toprule
\textbf{Models} & \multicolumn{3}{c}{\textbf{LLM Ensemble}} & \multicolumn{3}{c}{\textbf{Expert Crowdsourced}} & \multicolumn{3}{c}{\textbf{\rns Guided}}\\
\cmidrule(lr){2-4} \cmidrule(lr){5-7} \cmidrule(lr){8-10} 
 & \textbf{ Faithful.} & \textbf{Compre.} & \textbf{SerenCov} & \textbf{Faithful.} & \textbf{Compre.} & \textbf{SerenCov} & \textbf{Faithful.} & \textbf{Compre.} & \textbf{SerenCov} \\
\midrule
\
\dsh & 2.283 & 3.341 & 0.101 & 2.306 & 3.340 & 0.100 & 2.253 & 3.326 & 0.106 \\
\midrule
\tlal & \underline{2.519} & \textbf{3.842} & 0.070 & \underline{2.553} & \textbf{3.853} & 0.068 & \underline{2.531} & \textbf{3.829} & 0.075 \\
\tdsl & \textbf{2.573} & 2.206 & 0.223 & \textbf{2.572} & 2.238 & 0.204 & \textbf{2.582} & 2.202 & 0.217 \\
\tqwl & 2.024 & 2.683 & 0.153 & 2.093 & 2.715 & 0.152 & 2.114 & 2.719 & 0.155 \\
\midrule
\tmxm & 2.271 & 2.963 & \textbf{0.642} & 2.272 & 2.958 & \textbf{0.610} & 2.347 & 2.924 & \textbf{0.632} \\
\tqwm & 2.243 & 2.929 & 0.148 & 2.255 & 2.910 & 0.146 & 2.260 & 2.886 & 0.152 \\
\midrule
\tgmm & 2.365 & \underline{3.410} & 0.088 & 2.381 & \underline{3.439} & 0.084 & 2.385 & \underline{3.415} & 0.089 \\
\tmss & 2.114 & 3.016 & 0.141 & 2.114 & 3.048 & 0.136 & 2.134 & 3.049 & 0.141 \\
\midrule
\tqws & 1.920 & 1.817 & \underline{0.592} & 1.900 & 1.848 & \underline{0.580} & 1.955 & 1.832 & \underline{0.593} \\
\bottomrule
\end{tabular}
\end{adjustbox}
\begin{adjustbox}{max width=\textwidth}
\begin{tabular}{lccccccccc}
\toprule
\textbf{Models} & \multicolumn{3}{c}{\textbf{LLM Ensemble}} & \multicolumn{3}{c}{\textbf{Expert Crowdsourced}} & \multicolumn{3}{c}{\textbf{\rns Guided}}\\
\cmidrule(lr){2-4} \cmidrule(lr){5-7} \cmidrule(lr){8-10} 
 & \textbf{Relevance} & \textbf{TypeMatch} & \textbf{SerenHit} & \textbf{Relevance} & \textbf{TypeMatch} & \textbf{SerenHit} & \textbf{Relevance} & \textbf{TypeMatch} & \textbf{SerenHit} \\
\midrule
\dsh & 2.436 & 0.482 & \underline{0.048} & 2.494 & 0.462 & 0.061 & 2.538 & 0.463 & 0.077 \\
\hspace{2ex}$\hookrightarrow$ w.o. summary & 2.447 & 0.482 & \textbf{0.050} & 2.482 & 0.463 & \textbf{0.095} & 2.510 & 0.468 & \textbf{0.134} \\
\midrule
\tlal & \underline{2.537} & \underline{0.502} & 0.046 & \underline{2.559} & \textbf{0.483} & 0.067 & \underline{2.594} & \underline{0.478} & 0.106 \\
\hspace{2ex}$\hookrightarrow$ w.o. summary & \textbf{2.544} & \textbf{0.505} & 0.043 & \textbf{2.565} & \underline{0.478} & \underline{0.086} & \textbf{2.630} & \textbf{0.483} & \underline{0.127} \\
\tdsl & 1.935 & 0.424 & 0.030 & 2.000 & 0.409 & 0.034 & 2.033 & 0.418 & 0.049 \\
\hspace{2ex}$\hookrightarrow$ w.o. summary & 1.972 & 0.438 & 0.035 & 1.987 & 0.413 & 0.037 & 2.052 & 0.419 & 0.053 \\
\tqwl & 2.264 & 0.415 & 0.023 & 2.345 & 0.406 & 0.041 & 2.405 & 0.400 & 0.059 \\
\hspace{2ex}$\hookrightarrow$ w.o. summary & 2.269 & 0.428 & 0.028 & 2.337 & 0.416 & 0.050 & 2.409 & 0.412 & 0.070 \\
\midrule
\tmxm & 1.947 & 0.256 & 0.010 & 2.033 & 0.254 & 0.015 & 2.013 & 0.230 & 0.024 \\
\hspace{2ex}$\hookrightarrow$ w.o. summary & 2.158 & 0.324 & 0.016 & 2.250 & 0.312 & 0.022 & 2.220 & 0.306 & 0.042 \\
\tqwm & 2.294 & 0.441 & 0.036 & 2.331 & 0.426 & 0.045 & 2.378 & 0.429 & 0.065 \\
\hspace{2ex}$\hookrightarrow$ w.o. summary & 2.304 & 0.453 & 0.037 & 2.328 & 0.431 & 0.068 & 2.390 & 0.438 & 0.105 \\
\midrule
\tgmm & 2.357 & 0.450 & 0.033 & 2.379 & 0.414 & 0.057 & 2.443 & 0.431 & 0.080 \\
\hspace{2ex}$\hookrightarrow$ w.o. summary & 2.343 & 0.448 & 0.032 & 2.376 & 0.412 & 0.054 & 2.425 & 0.402 & 0.081 \\
\tmss & 1.855 & 0.195 & 0.008 & 1.959 & 0.184 & 0.016 & 2.005 & 0.185 & 0.026 \\
\hspace{2ex}$\hookrightarrow$ w.o. summary & 1.903 & 0.212 & 0.011 & 1.962 & 0.204 & 0.023 & 2.006 & 0.213 & 0.035 \\
\midrule
\tqws & 1.636 & 0.221 & 0.022 & 1.721 & 0.229 & 0.026 & 1.708 & 0.215 & 0.041 \\
\hspace{2ex}$\hookrightarrow$ w.o. summary & 1.487 & 0.160 & 0.018 & 1.550 & 0.175 & 0.018 & 1.547 & 0.158 & 0.027 \\

\bottomrule
\end{tabular}
\end{adjustbox}
\caption{Subgraph Reasoning ($T2$, upper), Serendipity Exploration ($T3$, lower), with Best scores \textbf{bolded}, 2nd best \underline{underlined}}
\vspace{-1.7ex}
\label{table:task3}
\end{table*}
\end{small}

\subsection{Task Analysis}
\label{sec:exp:tasks}

We next analyze experimental results task-by-task.

\stitle{Task 1: Knowledge Retrieval}. 
The results in Table~\ref{table:task1} show that larger models (e.g., \dsh, \gpt) 
consistently excel in simpler one-hop retrieval (F1 $\approx$ $78\%$), 
yet both exhibit performance degradation for more complex multi-hop queries (F1 drops to $<$ $10\%$ for queries with $3+$ hops). 
Smaller models are less accurate in coping with both simpler and more complex queries, reflecting limitations in reasoning depth and broader coverage of the biomedical context.
Notably, the two 70B models (\lal, \dsl) achieve better performances, 
which may be due to their more up-to-date training datasets. 

\stitle{Task 2: Subgraph Reasoning}.
In Table~\ref{table:task3} (upper), Mixtral-8×7B achieves (surprisingly) high Serendipity Coverage ($60\%+$) despite moderate scores in Faithfulness and Comprehensiveness ($2$-$3$ out of $5$). This interestingly indicates that summarization approaches yield broader serendipitous path coverage but risk factual inaccuracies. 
In contrast, larger models (\eg \lal) achieve higher Faithfulness and Comprehensiveness but lower ``SerenCov'', suggesting a consistent trade-off that their richer pre-trained knowledge produces more precise, yet narrower summaries. 

\stitle{Task 3: Serendipity Exploration}.
The rows labeled "w.o. summary" evaluate performance without subgraph summaries, isolating the effect of providing chain-of-thought guidance. For almost all models, removing the summary improved performance on all three metrics. One possible reason for this is that the model may introduce hallucinations during the summary process, which can influence the exploration path, as proven by Table~\ref{table:task3} (upper), many models did not achieve the desired score in subgraph reasoning.

\subsection{In-Depth Discussion}
\label{sec:exp:depth}

\stitle{Model scale vs.~Serendipity.}
As shown in the tables, larger models generally perform better in retrieval and exploration tasks. However, for subgraph summarization and reasoning (denoted as $T2$), there is significant variance and no obvious correlation with model size. This may suggest that retrieval and exploration benefit more from the model's inherent knowledge, which larger models excel at, while summarization and reasoning do not follow the same trend.

\begin{figure}[tb!]
\centerline{\includegraphics[width =0.47\textwidth]{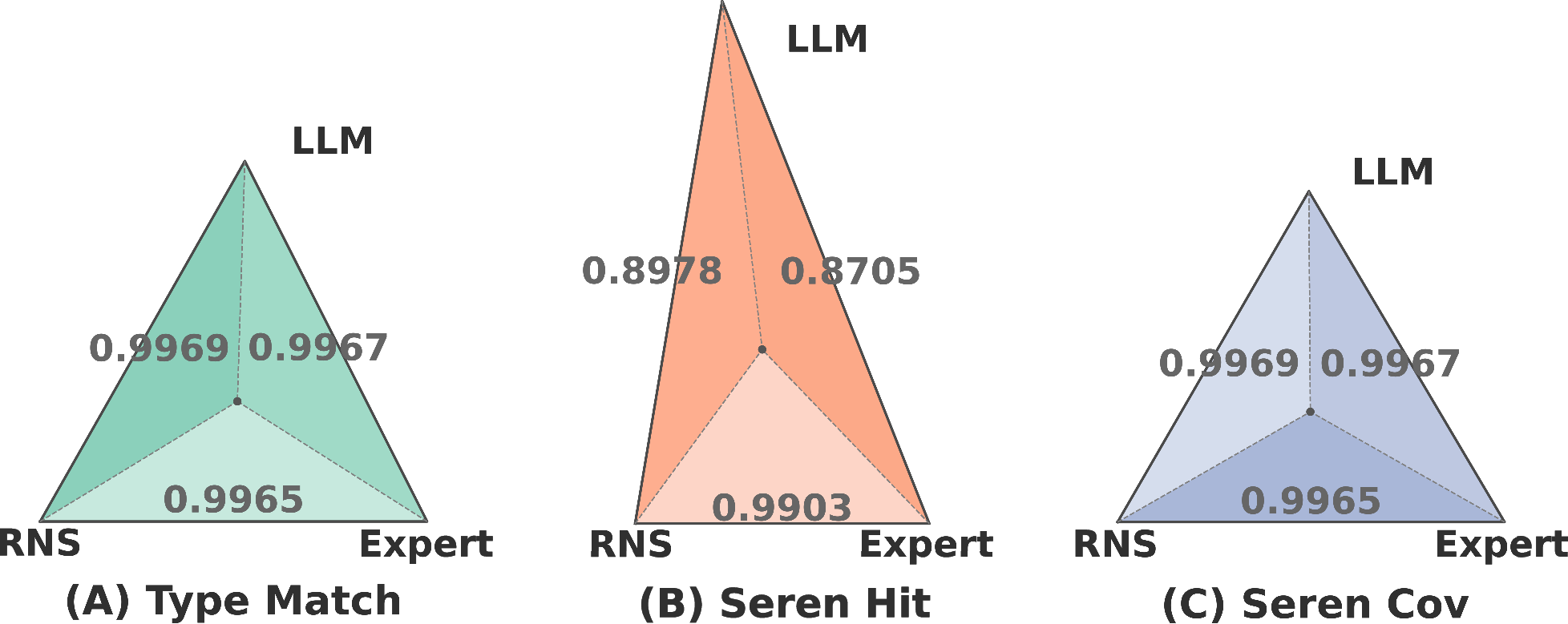}}
\centering
\caption{Correlation of Metrics Across Partition Strategies}
\label{fig:triangle_corr}
\vspace{-2ex}
\end{figure}

\stitle{Partition Sensitivity.}
Fig.~\ref{fig:triangle_corr} displays triangle plots of Pearson Correlations for TypeMatch, SerenCov, and SerenHit, with each triangle representing one metric. The corners denote three types of partitions, and edge weights indicate correlation scores—shorter distances refer to stronger correlations.
Our analysis shows that all partitions have positive correlations across all metrics, with scores above $85\%$. Notably, the expert and \rns-guided partitions reached around $99\%$ on all cases, highlighting the robustness of our partition strategies and the reliability of the proposed \rns measure.

\vspace{-1ex}

\stitle{No Single Winner.}
We found that no model constantly excels its peers across all metrics for each task. For instance, while Model \dsl performs excellently in retrieval, it shows only moderate performance in reasoning and poor results in exploration; \lal is more versatile but still struggles to address metrics from all perspectives.
To achieve balanced and serendipitous discovery, involving multiple models, such as multi-agent systems or a mixture of experts (MoE) strategy, may be beneficial.

We provide additional results and analysis in
Appendix~\ref{sec:app:exp}.

%% file: sections/conclusion.tex
\section{Conclusion}
\label{sec:conclusion}

We introduced \serenqa, an evaluation framework designed to assess 
LLMs' ability to discover serendipitous knowledge in scientific KGQA tasks.
We proposed an axiomatically justified serendipity measure integrating relevance, novelty, and surprise; 
and constructed a serendipity-aware benchmark tailored to the drug repurposing task. 
Additionally, we outlined a structured evaluation pipeline with three core tasks to assess LLM's ability on 
knowledge retrieval, subgraph reasoning, and serendipity exploration. 
Our experiments showed that frontier LLMs excel at basic knowledge retrieval, yet they often struggle with reasoning with more complex queries and answers for serendipity exploration, indicating great room and opportunities for improvement.

\section*{Ethical Statement}
In this study, we evaluated potential drug indications by analyzing biomedical relationships from ClinicalKG. Nevertheless, our approach does not consider factors critical to druggability, such as physicochemical properties.
 We used LLMs to identify serendipitous drug-disease associations that may suggest novel therapies. Their clinical effectiveness remains uncertain and must be validated through rigorous preclinical and clinical studies.
 


%% file: sections/appendix.tex
\onecolumn
\appendix

\sstab
This appendix contains the following content:

\textbf{A. Dataset Details}

- A.1 Dataset Construction

- A.2 Pattern Type

- A.3 Dataset Structure

- A.4 More Statistics

\textbf{B. Prompts}

- B.1 LLM Scoring Prompts

- B.2 Serendipity Exploration Prompts

- B.3 Pipeline Evaluation Prompts

\textbf{C. Further Analysis on \rns Metric}

- C.1 $k$-hop Conditional Probability Matrix

- C.2 Marginal Probability

\textbf{D. Details of Serensipity Exploration}

- D.1 Workflow and Logic

- D.2 Infrastructure

- D.3 Neighbor Scoring

\textbf{E. Experiment Details}

- E.1 Experiment Setting

- E.2 Additional Analysis

\vspace{2ex}
\hrule
\vspace{5ex}

\section{Dataset Details}
\label{sec:app:dataset}
\subsection{Dataset Construction}

\begin{figure*}[h!]
    \centering
    \includegraphics[width = 0.9\textwidth]{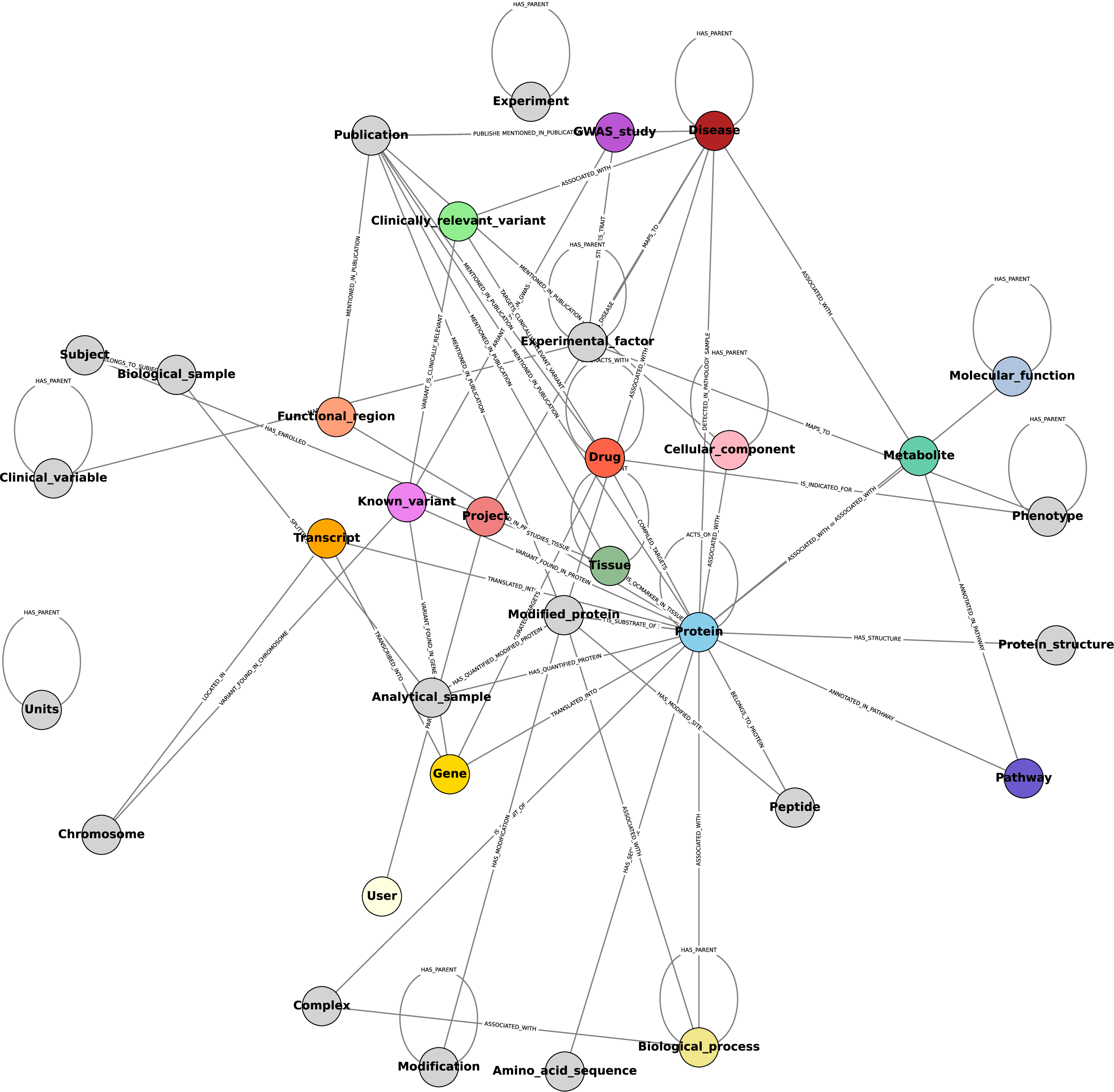}
    \centering
    \caption{Ontology of biomedical entities and relationships in the Clinical Knowledge Graph (CKG)}
    \label{fig:ontology}
\end{figure*}

We utilized the Clinical Knowledge Graph (CKG) \cite{Santos2022May} as the base knowledge graph to construct a benchmark question-answering dataset. A graph provides a structured organization of biomedical entities and their relationships, enabling systematic exploration and analysis of complex interactions. The CKG is built on curated public databases and literature-derived evidence, ensuring high-quality and biologically relevant information. Its comprehensive structure provides a robust foundation for generating diverse types of queries. The CKG encompasses $\sim$20 million nodes across 36 distinct types, including genes, proteins, diseases, drugs, pathways, anatomical entities, and other biological and clinical components, as shown in Fig~\ref{fig:ontology}. These nodes are interconnected by over 220 million edges spanning 47 different relationship types, capturing specific interactions and enabling detailed exploration of biomedical relationships, efficient data querying, and algorithmic analysis. Drug-phenotype relationships include edges such as ``has side effect'' and ``is indicated for,'' capturing drug effects and therapeutic indications. Gene-related relationships include ``variant found in gene'' and ``transcribed into,'' linking genetic variants to genes and transcripts, respectively, and highlighting structural and functional connections within the genome. Clinically relevant relationships, such as ``variant is clinically relevant'' and ``associated with,'' connect genetic variants to diseases. Additionally, drug-target interactions, captured by edges like “acts on” and “curated targets,” associate drugs with protein targets, offering insights into mechanisms of action and therapeutic potential.




To create the QA dataset, we extracted a subgraph of the CKG. Certain node and edge types, such as those related to users, units, experiments, projects, transcripts, and publications, were excluded to streamline the dataset and maintain focus on biologically significant relationships.

The current version of the dataset comprises 1,529 queries, with a focus on drug-disease associations, designed to evaluate the ability of large language models (LLMs) to identify serendipitous connections in the context of drug repurposing. Each query is annotated with relevant nodes, edges, and a target node, along with graph-specific metadata such as node and relationship types. We plan to continuously update and extend the dataset to include up to 5,000 queries, supporting a broader range of natural language processing tasks and a more comprehensive evaluation of LLM capabilities in biomedical reasoning.

The construction of the QA dataset involved several steps to optimize data retrieval and ensure its relevance to biomedical research. For one-hop and two-hop questions, the required data entries were extracted directly by querying the Neo4j database. For three-hop and intersection questions, given the computational demands of Neo4j queries and the large graphs, the relevant nodes and their one-hop neighborhoods were pre-extracted from the subgraph for more efficient processing.

To ensure the grammaticality, clarity, and biological relevance of the generated natural language questions, their phrasing was refined while preserving their original meanings. This involved programmatically extracting question patterns, retaining only the node types, and restructuring them into biologically meaningful and oncology-focused templates. 

\subsection{Pattern Type}
\label{sec:pattern}

The QA dataset includes one-hop, two-hop, three-hop, and intersection questions, designed to probe varying levels of complexity within the graph. Each type of question is defined by specific patterns, as described below:

\begin{enumerate}
    \item \textbf{One-hop questions:} These questions explore direct relationships between two entities connected by a single edge. They can be further categorized into two types:
    \begin{itemize}
        \item \textit{Type 1.1:} Questions that retrieve entities of a specific type (\{target\_type\}) connected to a source entity (\{source\_name\}) through a given relationship (\{relationship\}). For example, ``List the \{target\_type\}s that \{relationship\} by \{source\_name\}.'' and  "What \{source\_type\}s  \{relationship\} \{target\_name\}?" 
        \item \textit{Type 1.2:} Questions that identify the source entities (\{source\_type\}) connected to a specific target entity (\{target\_name\}) via a given relationship. For example, ``What \{source\_type\}s \{relationship\} \{target\_name\}?''
    \end{itemize}

    \item \textbf{Two-hop questions:} These questions traverse two edges, connecting a source entity to a final entity via an intermediate entity. Two patterns are defined:
    \begin{itemize}
        \item \textit{Type 2.1:} Questions that link the source entity (\{source\_name\}) to the final entity (\{final\_type\}) through an intermediate entity (\{mid\_name\}) and two relationships (\{relationship1\} and \{relationship2\}). For example, ``Which \{final\_type\} is \{relationship2\} by \{mid\_name\} that \{source\_name\} \{relationship1\}?''
    \end{itemize}

    \item \textbf{Three-hop questions:} These questions traverse three edges, uncovering chains of relationships across multiple intermediate entities. The questions explore how a source entity (\{entity1\}) connects to a final entity (\{entity4\}) through a sequence of intermediate entities (\{entity2\} and \{entity3\}). For example:
    \begin{itemize}
        \item \textit{Type 3.1:} Questions that trace a sequential path, e.g., ``Which \{entity4\_type\} is \{relationship3\} by \{entity3\_name\} that \{relationship2\} \{entity2\_name\} which \{relationship1\} \{entity1\_name\}?''
        \item \textit{Type 3.2:} Questions that incorporate hierarchical relationships, e.g., ``Which \{entity1\_type\} 
         \{relationship1\} \{entity2\_name\}, which \{relationship2\} \{entity3\_name\}, and \{relationship3\} \{entity4\_name\}?"
        \item \textit{Type 3.3:} Questions that branch into multiple connections, e.g., ``Which \{entity1\_type\} \{relationship1\} \{entity2\_name\} that \{relationship2\} \{entity3\_name\} and \{relationship3\} \{entity4\_name\}?''
    \end{itemize}

    \item \textbf{Intersection questions:} These focus on entities or sets of entities sharing multiple relationships with others. The goal is to identify overlapping connections across different paths within the graph. For example:
    \begin{itemize}
        \item \textit{Type 4.1:} Basic intersections, e.g., ``List \{entity1\_type\} that \{relationship1\} \{entity2\_name\} and \{relationship2\} \{entity3\_name\},'' which identify entities linked to two distinct targets through different relationships.
        \item \textit{Type 4.2:} Multi-way intersections,  such as ``List \{entity1\_type\} that \{relationship1\} \{entity2\_name\}, \{relationship2\} \{entity3\_name\}, and \{relationship3\} \{entity4\_name\},'' which extend to three overlapping connections.
        \item \textit{Type 4.3:} Compound intersections that involve cyclic patterns like ``Find all \{entity4\} that \{relationship43\} \{entity3\} and \{relationship42\} \{entity2\}, and also find the \{entity1\} that \{relationship13\} \{entity3\} and \{relationship12\} \{entity2\},'' in which entity2 and entity3 are connected to entity1 and entity4 through different links.
    \end{itemize}
\end{enumerate}

\subsection{Dataset Structure}
\label{sec:app:template}
Here, we introduce the structure of our drug repurposing benchmark, which supports both standard knowledge graph KGQA tasks and serendipity-aware evaluations.

As shown in the example below, each item in the QA dataset designed for standard KGQA tasks is standardized and configured to include expert-verified, scientifically meaningful NL queries, along with a structured Cypher query. Each query entry contains key components such as nodes, node types, and relationships, as well as a grounded and validated answer set. 
\begin{lstlisting}
 {
    "qid": 800,
    "question": "Which proteins are associated with dilated cardiomyopathy 1DD and function as subunits of the NOS3-HSP90 complex induced by VEGF?",
    "answer": [
      {
        "answer_type": "Entity",
        "answer_argument": "P29474",
        "entity_name": "NOS3",
      {
        "answer_type": "Entity",
        "answer_argument": "P07900",
        "entity_name": "HSP90AA1",
      }
    ],
    "function": "none",
    "commonness": 0.0,
    "num_node": 3,
    "num_edge": 2,
    "graph_query": {
      "nodes": [
        {
          "nid": 0,
          "node_type": "class",
          "id": "Protein",
          "class": "Protein",
          "friendly_name": "Protein",
          "question_node": 1,
          "function": "none"
        },
        {
          "nid": 1,
          "node_type": "entity",
          "id": "DOID:0110447",
          "class": "Disease",
          "friendly_name": "dilated cardiomyopathy 1DD",
          "question_node": 0,
          "function": "none"
        },
        {
          "nid": 2,
          "node_type": "entity",
          "id": "5716",
          "class": "Complex",
          "friendly_name": "NOS3-HSP90 complex, VEGF induced",
          "question_node": 0,
          "function": "none"
        }
      ],
      "edges": [
        {
          "start": 0,
          "end": 1,
          "relation": "Protein.Disease",
          "friendly_name": "ASSOCIATED_WITH"
        },
        {
          "start": 0,
          "end": 2,
          "relation": "Protein.Complex",
          "friendly_name": "IS_SUBUNIT_OF"
        }
      ]
    },
    "pattern_type": 9,
    "category": "genetic disease:autosomal genetic disease",
    "cypher": "MATCH (n0:Protein)\nMATCH (n1:Disease {name: \"dilated cardiomyopathy 1DD\"})\nMATCH (n2:Complex {name: \"NOS3-HSP90 complex, VEGF induced\"})\nMATCH (n0)-[:ASSOCIATED_WITH]->(n1)\nMATCH (n0)-[:IS_SUBUNIT_OF]->(n2)\n    RETURN\n    COLLECT(DISTINCT {id: n0.id, name: n0.name}) AS n0_targets"
  }
\end{lstlisting}

Below, we present the structure of the benchmark dataset designed to support serendipity-aware evaluation. The complete set of candidate answers obtained from the graph is partitioned into an existing set and a serendipity set using three distinct partitioning strategies detailed in Section 4.2.
We also provide the ground truth path from the existing set to the serendipity set for each query for possible training use.

\begin{lstlisting}
{
    "qid": 800,
    "question": "Which proteins are associated with dilated cardiomyopathy 1DD and function as subunits of the NOS3-HSP90 complex induced by VEGF?",
    "llm": {
        "serendipity_set": {
            "list": [
                "P29474"
            ],
            "description": null
        },
        "explore_queries": {
            "paths": [
                "P07900--COMPILED_INTERACTS_WITH--NOS2:Protein--BELONGS_TO_PROTEIN--None:Peptide--BELONGS_TO_PROTEIN--P29474",
                "P07900--ACTS_ON--NOS2:Protein--BELONGS_TO_PROTEIN--None:Peptide--BELONGS_TO_PROTEIN--P29474",
                "P07900--COMPILED_INTERACTS_WITH--NOS1:Protein--BELONGS_TO_PROTEIN--None:Peptide--BELONGS_TO_PROTEIN--P29474"
            ],
            "questions": [
                "Which proteins, interacting with NOS isoforms and belonging to the same protein complex as P07900, are involved in related molecular pathways?"
            ]
        },
        "partition": "test",
        "exact_matches": {
            "list": [
                "P07900"
            ]
        }
    },
    "sscore": {
        "serendipity_set": {
            "list": [
                "P07900"
            ],
            "description": null
        },
        "explore_queries": {
            "paths": [
                "P29474--ASSOCIATED_WITH--protein serine/threonine phosphatase complex:Cellular_component--HAS_PARENT--protein phosphatase type 2A complex:Cellular_component--ASSOCIATED_WITH--P07900",
                "P29474--COMPILED_INTERACTS_WITH--MAP2K1:Protein--ASSOCIATED_WITH--protein phosphatase type 2A complex:Cellular_component--ASSOCIATED_WITH--P07900",
                "P29474--COMPILED_INTERACTS_WITH--PPP2CA:Protein--ASSOCIATED_WITH--protein phosphatase type 2A complex:Cellular_component--ASSOCIATED_WITH--P07900"
            ],
            "questions": []
        },
        "partition": "test",
        "exact_matches": {
            "list": [
                "P29474"
            ]
        }
    },
    "expert": {
        "serendipity_set": {
            "list": [
                "P29474"
            ],
            "description": null
        },
        "explore_queries": {
            "paths": [
                "P07900--COMPILED_INTERACTS_WITH--NOS2:Protein--BELONGS_TO_PROTEIN--None:Peptide--BELONGS_TO_PROTEIN--P29474",
                "P07900--ACTS_ON--NOS2:Protein--BELONGS_TO_PROTEIN--None:Peptide--BELONGS_TO_PROTEIN--P29474",
                "P07900--COMPILED_INTERACTS_WITH--NOS1:Protein--BELONGS_TO_PROTEIN--None:Peptide--BELONGS_TO_PROTEIN--P29474"
            ],
            "questions": []
        },
        "partition": "test",
        "exact_matches": {
            "list": [
                "P07900"
            ],
            "description": null
        }
    }
}
\end{lstlisting}

\subsection{More Statistics}
\label{sec:app:stat}
The table below shows the composition of the KGQA dataset and the distribution of question pattern types among the 1,529 queries related to drug repurposing. The patterns for individual types are detailed in Appendix A.2.

\begin{table}[ht]
\centering
\begin{tabular}{l r}
\toprule
\textbf{Pattern Type} & \textbf{Number of Entries} \\
\midrule
\texttt{One hop Type 1.1}          & 152 \\
\texttt{One hop Type 1.2}  & 84  \\
\midrule
\texttt{Two hop Type 2.1}              & 152 \\
\midrule
\texttt{Three hop Type 3.1}            & 62  \\
\texttt{Three hop Type 3.2}            & 113 \\
\texttt{Three hop Type 3.3}            & 237 \\
\midrule
\texttt{Intersection Type 4.1}          & 263 \\
\texttt{Intersection Type 4.2}          & 455 \\
\texttt{Intersection Type 4.3}          & 11  \\
\bottomrule
\end{tabular}
\caption{Distribution of entries among different query pattern types.}
\label{tab:drug_disease_distribution}
\end{table}

\section{Prompts}
\label{sec:prompt}

\subsection{LLM Scoring Prompts}

\begin{lstlisting}[basicstyle=\small\ttfamily,breaklines=true,frame=single]
You are an expert evaluator specializing in drug discovery. Your task is to evaluate the **serendipity** of each answer in a provided list of answers, where all answers are derived from a knowledge base and are correct. Use your expertise and internal knowledge to assign a **serendipity score** to each answer based on the following criteria:

- **Serendipity Score**: A score from 0 to 20, where:
  - 20 represents an answer that is highly novel, unexpected, or insightful in the context of the question.
  - 0 represents an answer that is correct but very obvious, common, or provides no novel insights.
  - Intermediate scores represent varying degrees of novelty and insight.

- **Evaluation Rules**:
  1. The serendipity score reflects the relative novelty and insightfulness of each answer within the context of the question and the provided list. The score should highlight the uniqueness and unexpected value of each answer.
  2. Assign a distinct score to each answer. Even if multiple answers have a similar level of serendipity, assign slightly different scores to reflect the subtle differences in their uniqueness.
  3. Evaluate each answer independently of its position in the list.
  4. Output only the scores for each answer in the same order as the input list, separated by commas. Do not include the answers themselves or any additional explanation in the output.

For example:
If the input list is:
Answer List: A, B, C

The output should be:
5, 7, 9
\end{lstlisting}

  \subsection{Serendipity Exploration Prompts}
  
  \stitle{B.2.1 Select Relation}
  
  \stitle{System Prompt:}
  
  \begin{lstlisting}[basicstyle=\small\ttfamily,breaklines=true,frame=single]
  Task Description:
  Given a starting entity node (e.g., Drug, Protein, Disease), select the top-m relation types (predicates) to follow for meaningful, potentially serendipitous exploration in a clinical biomedical knowledge graph.
    
  Goal:
  Construct 3-hop paths that are:
    - Biologically plausible (based on frequent patterns)
    - Serendipitous (novel yet valid hypotheses)
    - Mechanistically rich (e.g., involving Drug-Protein-Disease chains)
    
  Path Patterns:
  Common patterns include:
    - (ACTS_ON, COMPILED_INTERACTS_WITH, ACTS_ON)
    - (INTERACTS_WITH, ACTS_ON, ACTS_ON)
    - (COMPILED_INTERACTS_WITH, ASSOCIATED_WITH, ASSOCIATED_WITH)
    
  Frequently explored node types:
    Drug, Protein, Disease, Gene, Metabolite
    
  Useful relation types:
    - Curated/compiled: CURATED_INTERACTS_WITH, COMPILED_TARGETS
    - Functional/structural: HAS_SEQUENCE, BELONGS_TO_PROTEIN
    - Annotations: ANNOTATED_IN_PATHWAY, DETECTED_IN_PATHOLOGY_SAMPLE
    - Rare/high-value: IS_INDICATED_FOR, HAS_SIDE_EFFECT, TRANSLATED_INTO
    
  Prioritize 3-hop sequences that reflect biological mechanisms. Balance high-frequency paths (plausibility) with rare combinations (serendipity). Avoid trivial paths unless used creatively.
    
  Output Requirements:
    - Return a comma-separated list of relation type strings
    - Do not include commentary or explanation
    - Use only the relation types provided as input
    - Return fewer than m results if appropriate
    - Return nothing if no meaningful exploration exists
    
  Notes:
    - Prioritize biologically important nodes and plausible mechanistic chains
    - Follow the path patterns listed above when applicable
    
  Few-Shot multi-hop example:
  Question: Which genes are identified as targets of D-Aspartic Acid, which affects ASPA and is known to interact with GLUD1?
  Root: GRIN2A, GRIN2C
  Serendipity set: GRIN2B
  Explore paths:
    - GRIN2A-TRANSLATED_INTO-GRIN2A:Protein-COMPILED_INTERACTS_WITH-GRIN2B:Protein-TRANSLATED_INTO-GRIN2B
    - GRIN2A-TRANSLATED_INTO-GRIN2A:Protein-ACTS_ON-GRIN2B:Protein-TRANSLATED_INTO-GRIN2B
    - GRIN2A-CURATED_TARGETS-Mesoridazine:Drug-INTERACTS_WITH-Felbamate:Drug-CURATED_TARGETS-GRIN2B
    - GRIN2C-TRANSLATED_INTO-GRIN2C:Protein-COMPILED_INTERACTS_WITH-GRIN2B:Protein-TRANSLATED_INTO-GRIN2B
    - GRIN2C-TRANSLATED_INTO-GRIN2C:Protein-ACTS_ON-GRIN2B:Protein-TRANSLATED_INTO-GRIN2B
    - GRIN2C-TRANSLATED_INTO-GRIN2C:Protein-ACTS_ON-D-Serine:Drug-CURATED_TARGETS-GRIN2B
  \end{lstlisting}
  
  \stitle{User Rrompt:}
  
  \begin{lstlisting}[basicstyle=\small\ttfamily,breaklines=true,frame=single]
  Given node ID {frontier} at level {level}, recommend the top {m} relation types to explore from this node.

  Context:
  {contexts}

  Available relation types from this node:
  {relation_types}

  Return a comma-separated list of relation type names only.
  \end{lstlisting}

  \stitle{B.2.2 Select Nodes}
  
  
  \begin{lstlisting}[basicstyle=\small\ttfamily,breaklines=true,frame=single]
  Task Description:
  You have already selected the most relevant relation types for exploring the graph from a given node. Now, for each selected relation, a set of target nodes has been retrieved.

  Goal:
  Construct 3-hop paths that are:
      - Biologically plausible (based on frequent patterns)
      - Serendipitous (novel yet valid hypotheses)
      - Mechanistically rich (e.g., involving Drug-Protein-Disease chains)

  The setting is a biomedical question answered in drug discovery. Exploration starts from known entities (e.g., drugs, proteins, diseases) and aims to discover serendipitous connections through meaningful 3-hop paths.

  Path Patterns:
  Common patterns include:
      - (ACTS_ON, COMPILED_INTERACTS_WITH, ACTS_ON)
      - (INTERACTS_WITH, ACTS_ON, ACTS_ON)
      - (COMPILED_INTERACTS_WITH, ASSOCIATED_WITH, ASSOCIATED_WITH)

  Frequently explored node types:
      Drug, Protein, Disease, Gene, Metabolite

  Useful relation types:
      - Curated/compiled: CURATED_INTERACTS_WITH, COMPILED_TARGETS
      - Functional/structural: HAS_SEQUENCE, BELONGS_TO_PROTEIN
      - Annotations: ANNOTATED_IN_PATHWAY, DETECTED_IN_PATHOLOGY_SAMPLE
      - Rare/high-value: IS_INDICATED_FOR, HAS_SIDE_EFFECT, TRANSLATED_INTO

  Prioritize 3-hop sequences that reflect biological mechanisms. Balance high-frequency paths (plausibility) with rare combinations (serendipity). Avoid trivial paths unless used creatively.

  Output Requirements

  Return a comma-separated list of selected target_ids only.
      - Do not include headers, explanations, or formatting.
      - If no target is suitable, return nothing.

  Constraints
      - Select only from relation types and target nodes provided by the user.
      - Do not include the current frontier node in the output.
      - Do not revisit nodes marked as already visited.
      - If fewer than n targets are appropriate, return fewer.
      - If exploration is not meaningful, return nothing.
      - Follow the path patterns listed above where applicable.
  \end{lstlisting}


\stitle{B.2.3 Decide Whether to Continue}

  \stitle{System Prompt:}
  
  \begin{lstlisting}[basicstyle=\small\ttfamily,breaklines=true,frame=single]
  Task Description
  
  You are exploring a biomedical knowledge graph in the context of drug discovery, starting from known entities (e.g., drugs, proteins, diseases) and aiming to uncover deeper, potentially serendipitous connections.
  
  In the previous two steps, you selected the most relevant relation types and target nodes for expansion. Before continuing, you must now:
      1. Review the full path from the root node to the current node (3-hop away).
      2. Provide a summary of the path's biological context.
      3. Decide whether further exploration is justified.
  
  Each input path is represented as a key-value pair:
      - Key: the current (destination) node ID
      - Value: a comma-separated sequence of alternating (target_id, relation_type) tuples tracing the 3-hop path from the root.
  
Use this information and the user's question to assess whether the exploration is still on a plausible, meaningful track toward the question objective.

  Output Requirements
  
  Your output must follow exactly the format below:
      1. A natural-language summary (~200 words), describing:
          - Biological meaning of the paths
          - Patterns of entity types
          - Common or notable relation sequences
          - Any biologically relevant interpretations
      2. (blank line)
      3. Either YES or NO, indicating whether to continue expanding
      4. (blank line)
      5. A one-paragraph explanation justifying your decision
  
  Do not include any extra commentary, formatting, bullet points, or sections outside this structure.
  
  Notes
      - Only return NO if you are ABSOLUTELY CONFIDENT the path has deviated from any biologically plausible trajectory.
      - When in doubt, continue exploring (YES).
      - Base your judgment on whether the current node plausibly supports mechanistic or therapeutic insight relevant to the original question.
  \end{lstlisting}

  \stitle{User Prompt:}
  
  \begin{lstlisting}[basicstyle=\small\ttfamily,breaklines=true,frame=single]
  The original question is:
  {question}
  
  The root node of the beam search is:
  {root}
  
  Subgraph paths (from root to current node):
  {paths}
  \end{lstlisting}

  \stitle{B.2.4 Summarize Subgraph}
  
  \stitle{System Prompt:}
  
  \begin{lstlisting}[basicstyle=\small\ttfamily,breaklines=true,frame=single]
  You are an expert biomedical knowledge graph assistant. You have performed a beam search starting from a root node over a clinical biomedical knowledge graph, retrieving 1-hop, 2-hop, and 3-hop subgraphs.
  
  Output Requirement
  
  Provide a concise natural-language summary (~200 words) of the resulting subgraphs.
      - Mention as many specific biomedical terms (e.g., drugs, proteins, diseases, pathways) as possible.
      - Emphasize the types of entities and the patterns of relationships involved.
      - Focus on the biological meaning, mechanistic implications, or potential therapeutic relevance of the paths.
  
Do not include any formatting, headers, or commentary--only the summary text.

  \end{lstlisting}

  \stitle{User Prompt:}
  
  \begin{lstlisting}[basicstyle=\small\ttfamily,breaklines=true,frame=single]
  Root node ID:
  {root}
  
  Question:
  {question}
  
  Hop level:
  {level}
  
  Subgraph paths (from root to leaf nodes):
  {subgraph}
  \end{lstlisting}

  \subsection{Pipeline Evaluation Prompts}
  
  \stitle{B.3.1 Faithfulness Assessment}
  
  \stitle{System Prompt:}
  
  \begin{lstlisting}[basicstyle=\small\ttfamily,breaklines=true,frame=single]
  You are assisting a multi-stage research pipeline that explores a large biomedical
  knowledge graph.

  Pipeline stages
  **Exact-Match Retrieval** -- find entities that directly answer the user's question
     (these are the "root" nodes).
  **Serendipity Exploration** -- expand <=3 hops from the root to propose *new*,
     potentially surprising but biologically meaningful entities
     (the "exploration result" is captured by the **paths** and the **leaves**).
  **Hop-level Summaries** -- for readability, the pipeline auto-generates three short
     natural-language summaries:
       * *summary 1*  -> describes the 1-hop neighbourhood around the root
       * *summary 2*  -> describes the 2-hop sub-graph discovered next
       * *summary 3*  -> describes the 3-hop sub-graph plus any thematic insight

  You receive:
  --------------------------------------------------------
  * root            -- the starting entity ID (protein / drug)
  * question        -- original natural-language question
  * summary_1/2/3   -- auto-generated summaries of the 1-hop, 2-hop, and
                      3-hop neighbourhoods around the root
  * leaves          -- **all endpoint nodes in the explored sub-graph**
                      (may be 1-, 2-, or 3-hop away) -- each item is given
                      as  <node_id>(<node_type>)  e.g.  `P52209(Protein)`
  * paths           -- ground-truth triples, one per line, with types included:
                      head_id(head_type),relation_type,tail_id(tail_type)
  --------------------------------------------------------

  Task
  ====
  * First, read the sub-graph and understand every factual triple
    it contains.
  * Then, read the three hop-summaries in order (1-hop -> 3-hop).
  * "Faithfulness" here means: *How truthfully do the summaries reflect
    what is actually present in the graph, without inventing new entities,
    directions, or relations?*
    - Higher faithfulness -> few to no hallucinations or distortions.
    - Lower faithfulness -> noticeable fabrication, wrong direction,
      or missing key context.

  Using your best expert judgment of biomedical knowledge-graphs,
  assign a holistic integer score:

      5  - Completely faithful
      4  - Mostly faithful, only trivial wording drift
      3  - Mixed: some accurate, some questionable
      2  - Largely unfaithful, many doubtful claims
      1  - Almost entirely unfaithful / hallucinated

  Do **not** count tokens or sentences; rely on your overall sense of truthfulness.

  IMPORTANT: If the input is completely empty or contains no evaluable information whatsoever,
  return Score: 1. However, if there is ANY evaluable content, even if partial or limited,
  evaluate it based on the 1-5 scale above. Do not argue or explain if content is missing, 
  just assign the appropriate score and return the two required lines.

  Output format
  -------------
  Return **exactly** these two lines--nothing more, nothing less:

  Score: <INTEGER 1-5>
  #END
  \end{lstlisting}

  \stitle{User Prompt:}
  
  \begin{lstlisting}[basicstyle=\small\ttfamily,breaklines=true,frame=single]
  Root: {root}
  Question: {question}

  -- 1-Hop Summary --
  {summary_1}

  -- 2-Hop Summary --
  {summary_2}

  -- 3-Hop Summary --
  {summary_3}

  Leaf nodes: {leaves}

  Sub-graph (Triples):
  {paths}
  \end{lstlisting}

  \stitle{B.3.2 Comprehensiveness Assessment}
  
  \stitle{System Prompt:}
  
  \begin{lstlisting}[basicstyle=\small\ttfamily,breaklines=true,frame=single]
  You are assisting a multi-stage research pipeline that explores a large biomedical
  knowledge graph.

  Pipeline stages
  **Exact-Match Retrieval** -- find entities that directly answer the user's question
     (these are the "root" nodes).
  **Serendipity Exploration** -- expand <=3 hops from the root to propose *new*,
     potentially surprising but biologically meaningful entities
     (the "exploration result" is captured by the **paths** and the **leaves**).
  **Hop-level Summaries** -- for readability, the pipeline auto-generates three short
     natural-language summaries:
       * *summary 1*  -> describes the 1-hop neighbourhood around the root
       * *summary 2*  -> describes the 2-hop sub-graph discovered next
       * *summary 3*  -> describes the 3-hop sub-graph plus any thematic insight

  You receive:
  --------------------------------------------------------
  * root            -- the starting entity ID (protein / drug)
  * question        -- original natural-language question
  * summary_1/2/3   -- auto-generated summaries of the 1-hop, 2-hop, and
                      3-hop neighbourhoods around the root
  * leaves          -- **all endpoint nodes in the explored sub-graph**
                      (may be 1-, 2-, or 3-hop away) -- each item is given
                      as  <node_id>(<node_type>)  e.g.  `P52209(Protein)`
  * paths           -- ground-truth triples, one per line, with types included:
                      head_id(head_type),relation_type,tail_id(tail_type)
  --------------------------------------------------------

  Task
  ====
  * First, study the sub-graph so you grasp **every** entity and
    relation present within three hops of the root.
  * Then, read the three hop-summaries in order (1-hop -> 3-hop).
  * "Comprehensiveness" here means: *How thoroughly do the summaries cover
    the important entities, relations, and mechanistic chains in the graph--
    without ignoring major facts?*
    - Higher Comprehensiveness -> almost all salient triples or concepts appear.
    - Lower Comprehensiveness -> key relationships, nodes, or overall structure
      are missing or only vaguely hinted at.

  Using your best expert judgment (no counting rules), assign a holistic
  integer score:

      5  - Nearly everything important is covered
      4  - Most key content covered; minor omissions
      3  - About half of the important content represented
      2  - Many significant omissions; partial picture
      1  - Very little of the important content included

  Do **not** estimate by token length; base the score on your global sense of
  coverage and relevance.

  IMPORTANT: If the input is completely empty or contains no evaluable information whatsoever,
  return Score: 1. However, if there is ANY evaluable content, even if partial or limited,
  evaluate it based on the 1-5 scale above. Do not argue or explain if content is missing, 
  just assign the appropriate score and return the two required lines.

  Output format
  -------------
  Return **exactly** these two lines--nothing more, nothing less:

  Score: <INTEGER 1-5>
  #END
  \end{lstlisting}

  \stitle{User Prompt:}
  
  \begin{lstlisting}[basicstyle=\small\ttfamily,breaklines=true,frame=single]
  Root: {root}
  Question: {question}

  -- 1-Hop Summary --
  {summary_1}

  -- 2-Hop Summary --
  {summary_2}

  -- 3-Hop Summary --
  {summary_3}

  Leaf nodes: {leaves}

  Sub-graph (Triples):
  {paths}
  \end{lstlisting}

  \stitle{B.3.3 Relevance Assessment}
  
  \stitle{System Prompt:}
  
  \begin{lstlisting}[basicstyle=\small\ttfamily,breaklines=true,frame=single]
  You are assisting a multi-stage research pipeline that explores a large biomedical
  knowledge graph.

  Pipeline stages
  **Exact-Match Retrieval** -- find entities that directly answer the user's question
     (these are the "root" nodes).
  **Serendipity Exploration** -- expand <=3 hops from the root to propose *new*,
     potentially surprising but biologically meaningful entities (the "predicted
     serendipity set").  These are evaluated against a **ground-truth serendipity
     set** that was curated by domain experts.

  You are rating how well a *predicted* serendipity answer set aligns with a
  *ground-truth* serendipity answer set that has been manually verified by
  domain experts.

  Facts you MUST assume:
  * The ground-truth set is correct.
  * Each ground-truth entity has been verified to be "serendipitous" with
    respect to the current exact-match root (i.e., useful and non-obvious
    extensions beyond that root).

  How to judge "relevance"
  > Does each predicted entity belong to the same mechanistic pathway,
    disease context, pharmacological class, or molecular family implied by
    the ground-truth set?
  > Overlap in **type** (Protein, Drug, Disease, Phenotype...) is helpful but
    not sufficient--focus on functional or clinical relatedness.
  > Minor naming variants or isoforms of a ground-truth entity are acceptable.

  Scoring rubric (integer)
  5 - Every prediction is clearly relevant;
  4 - Most (~ 70-90 %) predictions are relevant; few marginal or tangential items  
  3 - Mixed: roughly half relevant, half off-topic or trivial
  2 - Only a small minority appear relevant; set is mostly noise
  1 - Predictions are unrelated, incorrect, or obviously random

  IMPORTANT: If the input is completely empty or contains no evaluable information at all,
  return Score: 1. However, if there is ANY evaluable content, even if partial or limited,
  evaluate it based on the 1-5 scale above. Do not argue or explain if content is missing, 
  just assign the appropriate score and return the two required lines.

  Output format
  -------------
  Return **exactly** these two lines--nothing more, nothing less:

  Score: <INTEGER 1-5>
  #END
  \end{lstlisting}

  \stitle{User Prompt:}
  
  \begin{lstlisting}[basicstyle=\small\ttfamily,breaklines=true,frame=single]
  Original question:
  {question}

  Ground-truth serendipity set (trusted):
  {gold_seren}

  Predicted serendipity set (to be scored):
  {pred_seren}

  Exact-match root entity: {root}

  Hop-level summaries:
    * Level-1 -> {summary_1}
    * Level-2 -> {summary_2}
    * Level-3 -> {summary_3}

  Contextual graph paths:
  {paths}
  \end{lstlisting}

\section{Further Analysis on \rns Metric}
\label{sec:app:score}

\subsection{$k$-hop Conditional Probability Matrix}

\subsubsection{Properties Verification}

As defined in Sec.~\ref{sec:modeling}, the $k$-hop conditional probability matrix $P_k$ is 
\begin{small}
computed as:
$$
P_k = \sum_{h=1}^{k} \alpha_h P_1^h, 
\quad  \alpha_h = \frac{h}{\sum_{h=1}^{k} h} 
$$ 
\end{small}
We next prove that $P_k$ remains a valid transition probability matrix by verifying two essential properties explicitly:
\tbi
\item \textit{Non-negativity}: $(P_k)_{ij} \geq 0$ for all $(i, j)$,
\item \textit{Row-Stochastic Property}: $\sum_{j} (P_k)_{ij} = 1$  for all $i$.
\ei

\eetitle{Non-negativity.}
Since $P_1$ is directly derived from the adjacency matrix and row-normalized, all its elements are non-negative. 
Consequently, any power $P_1^h$ (for $h \geq 1$) is also non-negative, as it results from repeated multiplications of non-negative matrices. 
Furthermore, the weight coefficients $a_h$ are clearly positive by definition. Therefore, the linear combination $P_k = \sum_{h=1}^{k} \alpha_h P_1^h$ consists only of non-negative terms, ensuring:
$(P_k)_{ij} \geq 0, \forall (i, j)$.

\eetitle{Row-Stochastic Property.}
For $P_k$ to be a valid transition matrix, every row must sum exactly to one:
$$
\sum_{j} (P_k)_{ij} = 1, \quad \forall i
$$
We explicitly verify this condition: 
$$
\sum_{j} (P_k)_{ij} = \sum_{j} \sum_{h=1}^{k} \alpha_h (P_1^h)_{ij}
$$
Exchanging summation order (by linearity) yields:
$$
\sum_{j} (P_k)_{ij} = \sum_{h=1}^{k} \alpha_h \sum_{j} (P_1^h)_{ij}
$$
Since $P_1^h$ is a valid transition matrix, by definition, we have:
$$
\sum_{j} (P_1^h)_{ij} = 1, \quad \forall i, h
$$
Substituting the definition of $a_h$:
$$
\sum_{h=1}^{k} \alpha_h = \frac{1}{\sum_{h=1}^{k} h} \sum_{h=1}^{k} h = \frac{\sum_{h=1}^{k} h}{\sum_{h=1}^{k} h} = 1.
$$
Hence,
$$
\sum_{j} (P_k)_{ij} = 1, \forall i.
$$
Confirming that $P_k$ maintains row-stochasticity.

In summary, we've shown clearly that $P_k$ is both non-negative and row-stochastic. Therefore, the weighted multi-hop combination $P_k$ remains a valid transition probability matrix.

\subsubsection{Computation} 
To efficiently compute $P_k$, we apply a divide-and-conquer matrix multiplication approach based on Strassen's algorithm~\cite{strassen1969gaussian}. Specifically, the algorithm recursively divides each large $V \times V$ matrix into four sub-matrices of size $\frac{V}{2} \times \frac{V}{2}$. By strategically reusing these sub-matrix computations, Strassen's method reduces the number of necessary multiplications per recursion from the standard eight down to seven, thereby lowering the complexity significantly from the naive $\mathcal{O}(V^3)$ to approximately $\mathcal{O}(V^{\log_2 7}) \approx \mathcal{O}(V^{2.807})$. Moreover, parallelizing these recursive computations across $t$ processors further reduces the complexity to about $\mathcal{O}(V^{\log_2 7}/t)$. This ensures scalable and efficient computation of multi-hop conditional probability matrices, even for large-scale graphs.

\subsection{Marginal Probability}

We approximate the marginal probability computation via a PageRank-style damped iteration (Algorithm~\ref{alg:marginal-probability}).
For each iteration, 
\begin{itemize}
    \item Multiplying an $V \times V$ matrix $P_3^T$ by a vector $\mathbf{P}_t$ requires complexity $\mathcal{O}(V^2)$.
    \item Updating $\mathbf{P}_{t+1}$ is $\mathcal{O}(V)$, dominated by matrix-vector multiplication.
    \item Computing the difference $\|\mathbf{P}_{t+1}-\mathbf{P}_{t}\|_1$ takes $\mathcal{O}(V)$.
\end{itemize}
Hence, each iteration's complexity is dominated by the matrix-vector multiplication step, which is $\mathcal{O}(V^2)$.

\begin{algorithm}[tb]
\caption{Marginal Probability via PageRank-style Iteration}
\label{alg:marginal-probability}
\textbf{Input:} $P_3 \in \mathbb{R}^{V\times V}$, damping factor $\lambda$, tolerance $\epsilon$\\
\textbf{Output:} Marginal probability vector $\mathbf{P}\in\mathbb{R}^{V\times 1}$
\begin{algorithmic}[1]
\State Initialize $\mathbf{P}_0(i) := 1/V$, for all nodes $i$
\State $t := 0$
\While{$\text{diff} \geq \epsilon$}
    \State $\mathbf{P}_{t+1} := \lambda P_3^T \mathbf{P}_{t} + (1 - \lambda)\mathbf{P}_0$
    \State $\text{diff} := \|\mathbf{P}_{t+1}-\mathbf{P}_{t}\|_1$
    \State $t := t+1$
\EndWhile
\State \Return $\mathbf{P}_{t}$
\end{algorithmic}
\end{algorithm}

The error at iteration $t$ satisfies:
$
\|\mathbf{P}_t - \mathbf{P}_{t-1}\|_1 \leq c\lambda^t (0 < \lambda < 1)
$
for some constant $c$. Thus, convergence to within tolerance $\epsilon$ occurs after approximately:
$$
\lambda^t \approx \epsilon \quad \Rightarrow \quad t \approx \frac{\log(\epsilon)}{\log(\lambda)} = \mathcal{O}(\log V).
$$
This implies the total complexity to achieve convergence within $\epsilon$ is $\mathcal{O}(V^2 \log V)$.

\section{Details of Serensipity Exploration}
\label{sec:app:ppl}
\subsection{Workflow and Logic}
We designed a multi-stage beam search pipeline for structured knowledge graph exploration, 
as shown in Algorithm~\ref{alg:beam-explore},
where the expansion at each stage is guided by LLM. The pipeline explores neighborhoods of the root node recursively over the knowledge graph, while integrating external reasoning via multiple LLM interactions.

\begin{algorithm}[tb!]
\caption{LLM Enhanced Beam Explore}
\label{alg:beam-explore}

\textbf{Input:}

\begin{tabular}{@{}ll@{}}
$G = (V, E)$              & directed knowledge graph, \\
$n \in \mathbb{N}^+$      & beam width, \\
$m \in \mathbb{N}^+$      & max relation types per frontier node, \\
$k \in \mathbb{N}^+$      & max nodes to select per frontier node, \\
$h \in \mathbb{N}^+$      & beam depth, \\
$q \in \text{String}$     & natural language question, \\
$r_0 \in V$               & root node ID, \\
$context \in \{w, wo\}$   & context mode flag, \\
\end{tabular}

\textbf{Output:}

\begin{tabular}{@{}ll@{}}
$\Pi$          & node-to-path map from root, \\
$\Sigma$       & LLM-generated summaries per level, \\
\end{tabular}

\textbf{Definitions:}

\begin{tabular}{@{}ll@{}}
$\mathcal{V}$ & set of visited nodes, corresponds to \texttt{visited}, \\
$\mathbb{F}$ & current frontier nodes, corresponds to \texttt{frontier}, \\
$\mathbb{F}'$ & next frontier nodes, corresponds to \texttt{next\_frontier}, \\
$\mathcal{E}$ & set of candidate edges, corresponds to \texttt{candidates}, \\
$\mathcal{C}$ & LLM context buffer, corresponds to \texttt{context\_buffer}, \\
$d^\ast$  & leaf depth reached, corresponds to \texttt{leaf\_depth}, \\
\end{tabular}\\

\begin{algorithmic}[1]

\State \textbf{Initialize:}\\
\hspace{1.5em}$\mathcal{V} := \emptyset$ \\
\hspace{1.5em}$\Pi[r_0] := []$ \\
\hspace{1.5em}$\mathbb{F} := \{r_0\}$ \\
\hspace{1.5em}$\mathcal{C} := \emptyset$ \\
\hspace{1.5em}$d^\ast := 1$\\

\For{$\text{level} = 1$ \textbf{to} $h$}
    \State \textbf{set} $\mathbb{F}' := \emptyset$;\; $\mathcal{E} := \emptyset$
    \For{\textbf{each} node $u \in \mathbb{F}$}
        \State $R := \{r \mid (u, r, v) \in E\}$ 
        \If{$R = \emptyset$} \textbf{continue} \EndIf
        \State $R^m := \texttt{LLM\_SelectRelations}(q, u, R, m, \text{level}, \mathcal{C})$
        \State $C := \{(u, r, v) \in E \mid r \in R^m\}$
        \State $\mathcal{E} := \mathcal{E} \cup C$
    \EndFor
    \If{$\mathcal{E} = \emptyset$} \textbf{break} \EndIf
    \For{\textbf{each} edge $e \in \mathcal{E}$}
        \State \textbf{try} $e.\text{score} := \texttt{Score}(e)$ \textbf{except} $e.\text{score} := -1$
    \EndFor
    \If{$\forall e \in \mathcal{E},\; e.\text{score} = -1$}
        \State $\mathcal{E} := \texttt{UniformSample}(\mathcal{E}, \min(k, |\mathcal{E}|))$
    \Else
        \State $\mathcal{E} := \texttt{FilterTopKByScore}(\mathcal{E}, k)$
    \EndIf
    \State $V^d := \texttt{LLM\_SelectNodes}(q, u, \mathcal{E}, n, \text{level}, \mathcal{C})$
    \For{\textbf{each} $(u, r, v) \in \mathcal{E}$ \textbf{where} $v \in V^d$}
        \State $\Pi[v] := \Pi[u] \mathbin\Vert (u, r, v)$
        \State $\mathbb{F}' := \mathbb{F}' \cup \{v\}$
    \EndFor
    \State $\mathcal{V} := \mathcal{V} \cup V^d$
    \State $\mathcal{C}[\text{level}] := \texttt{LLM\_DescribePaths}(q, r_0, \Pi)$
    \State $\text{decision} := \texttt{LLM\_ShouldContinue}(q, r_0, \Pi)$
    \If{$\text{decision} = \texttt{no}$} \textbf{break} \EndIf
    \State $\mathbb{F} := \mathbb{F}'$
    \State $d^\ast := \text{level}$
\EndFor
\For{$l = 1$ \textbf{to} $d^\ast$}
    \State $\Sigma[l] := \texttt{LLM\_Summarize}(\Pi \text{ of depth } l)$
\EndFor
\State \textbf{return} $(\Pi, \Sigma)$
\end{algorithmic}
\end{algorithm}

\subsection{Infrastructure}
To support large-scale knowledge graph exploration, we constructed a lightweight compute-storage cluster on AWS, designed for high-throughput, low-latency edge retrieval and efficient task scheduling.
The cluster consists of two tiers of instances: compute nodes (5 * r6a.24xlarge) and storage nodes (5 * r6a.xlarge). Each compute node provides 96 vCPUs and 768 GiB of memory, serving as task executors that support large scale parallelism. Each storage node is configured as Redis servers through Docker container and acts as distributed read-only data backends for edge access.

To achieve high performance, we replaced Neo4j with a custom Redis-based edge storage scheme. The complete knowledge graph was exported from Neo4j and the key of each edge is encoded as (rel:\{source\_id\}:\{source\_type\}:\{relation\_type\}:\{target\_id\}:\{target\_type\}). The value stores metadata of relations and additional attributes for further use. The shift in query style allows us to improve query performance from 1000 QPS to tens of thousands of QPS. Each compute node interacts asynchronously with the storage node; empirically, the system supports a concurrency level of approximately 100 per compute node, enabling efficient exploration of multi-hop paths.

In order to facilitate our experimental process, we implemented an SSH-based compute cluster manager, responsible for task dispatching, resource allocation, permission control, environment setup, and declarative node specification. This infrastructure allows rapid iteration, cost-efficient experiments, and consistent resource management across multiple runs.

\subsection{Neighbor Scoring}
Some nodes have an extremely large number of neighbors as hub nodes. To effectively guide the LLM in exploring and reducing token usage, we design a scoring mechanism to reduce the size of nodes provided to the LLM.

We systematically extracted and quantified edge-level connection strength/confidence from a Neo4j-based clinical knowledge graph to support downstream analysis of biomedical associations. For relationship types such as ``ASSOCIATED\_WITH", ``COMPILED\_INTERACTS\_WITH", and ``ACTS\_ON", we directly extracted precomputed confidence scores. For other edge types, custom scoring functions were implemented based on domain-specific semantics. For instance, in the case of ``DETECTED\_IN\_PATHOLOGY\_SAMPLE", an expression score was derived using a weighted scheme based on categorical expression levels (e.g., high, medium, low, and not detected), while a prognostic score was computed using log-transformed p-values representing positive and negative survival associations. Both scores were then min-max normalized and aggregated to produce a final quantitative estimate reflecting the biomarker’s expression and clinical prognostic significance. This structured quantification enabled consistent interpretation and prioritization of heterogeneous relationship types within the graph.

\section{Experiment Details}
\label{sec:app:exp}

\subsection{Experiment Setting}

\stitle{Models:} We evaluated a wide range of state-of-the-art language models' using by evaluation setting:

  \begin{itemize}[leftmargin=*]
  \item \textit{Frontier Models}: \dsh, \gpt, \claude;
  
  \item \textit{Large Models ($\sim$70B)}: Llama-3.3-70B, DeepSeek-R1-70B, and Qwen-2.5-72B, \medl;
  
  \item \textit{Medium Models ($20$-$50$B)}: Mixtral-8x7B, \qwm, \dsm, \tgmm, \tmss

  \item \textit{Small Model}: Qwen-2.5-7B, \qws, \dss, \meds, \qwt, \dst
  \end{itemize}


\stitle{Metric Details}
We next detailed how we compute the metrics in subgraph reasoning and serendipity exploration tasks.

\eetitle{Subgraph Reasoning} All metrics are averaged across numbers of rational samples to give the final result (sum of metrics from rational samples/number of rational samples): 

(1) \textit{Faithfulness} (1-5, LLM-judged) - how truthfully do the summaries reflect what is actually present in the graph, without inventing new entities, directions, or relations?

(2) \textit{Comprehensiveness} (1-5, LLM-judged) - how thoroughly do the summaries cover the important entities, relations, and mechanistic chains in the graph?

(3) \textit{Serendipity Coverage} (0-1, code-based) -  fraction of serendipity paths where BOTH source and target node IDs are explicitly mentioned in the summary text. No LLM evaluation, just regex matching of node IDs.

\eetitle{Serendipity Exploration} All metrics are averaged across numbers of rational samples to give the final result (sum of metrics from rational samples/number of rational samples): 

(1) \textit{Relevance} (1-5, LLM-judged) - how well predicted serendipity entities align with the ground-truth serendipity set.

(2) \textit{TypeMatch} (0-1, code-based) - returns 1 if ANY predicted leaf has a type matching ground-truth types, 0 otherwise.

(3) \textit{SerenHit} (0-1, code-based) - returns $1$ if ANY predicted leaf is exactly matching ground-truth serendipity set (not just the type), $0$ otherwise.

\subsection{Additional Analysis}

We provide further analysis with supplementary figures to support and clarify key observations made in Section~\ref{sec:exp}.

\begin{figure*}[h!]
    \centering
    \includegraphics[width = \textwidth]{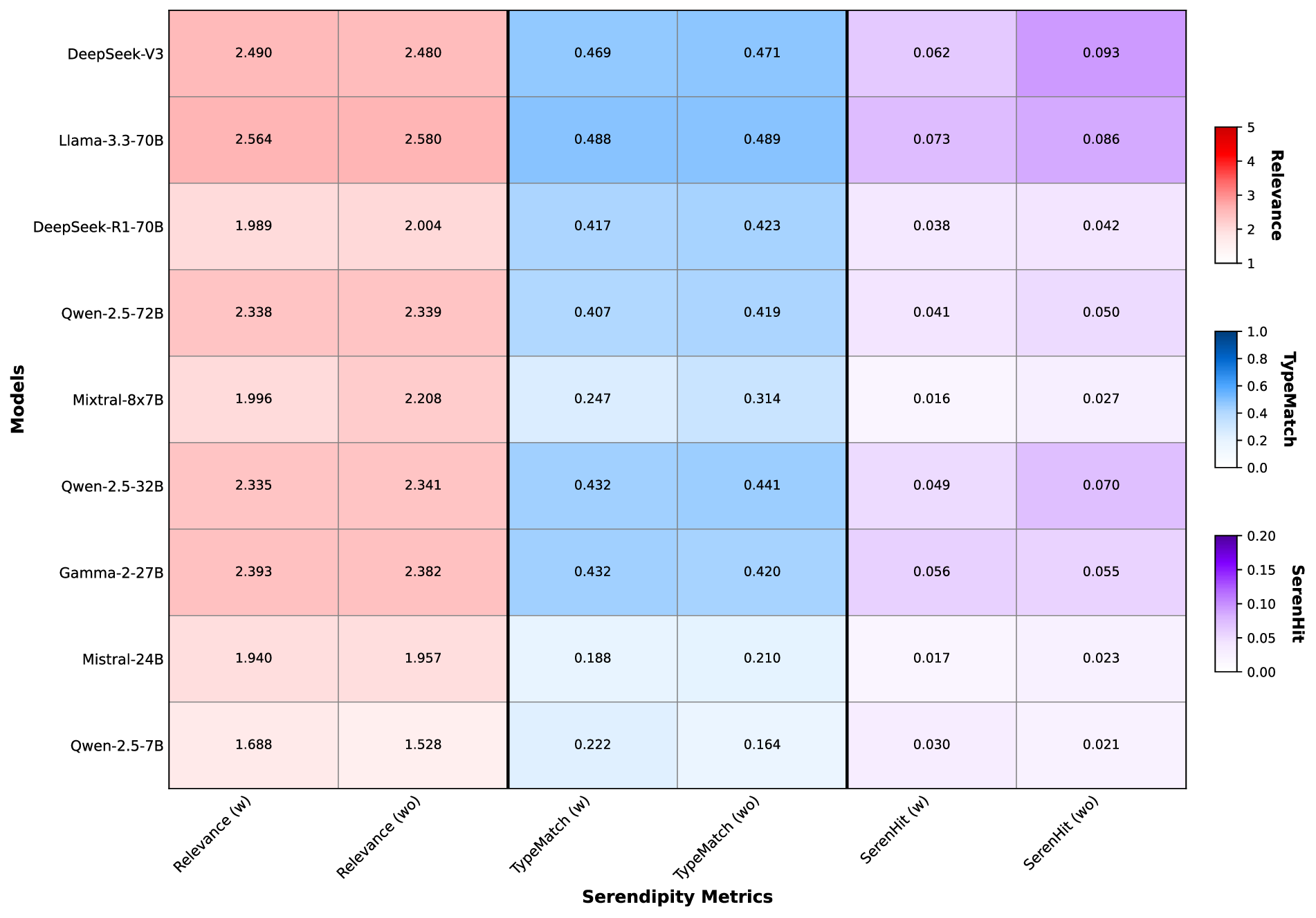}
    \centering
    \caption{Model scale vs. Serendipity Exploration Performance}
    \label{fig:heatmap}
\end{figure*}

\stitle{Model Scale vs. Serendipity Exploration} 
The heat-map shown in Fig.~\ref{fig:heatmap} analysis shows only a modest performance gain as model size increases from smaller (~$7$B) to larger (~$70$B) checkpoints. Relevance scores gradually improve, but TypeMatch and SerenHit increase inconsistently, with SerenHit remaining relatively low ($<$ 0.10). Although model scale contributes positively, larger parameters alone are insufficient to reliably achieve precise serendipitous discovery.

\begin{figure*}[h!]
    \centering
    \includegraphics[width = 0.85\textwidth]{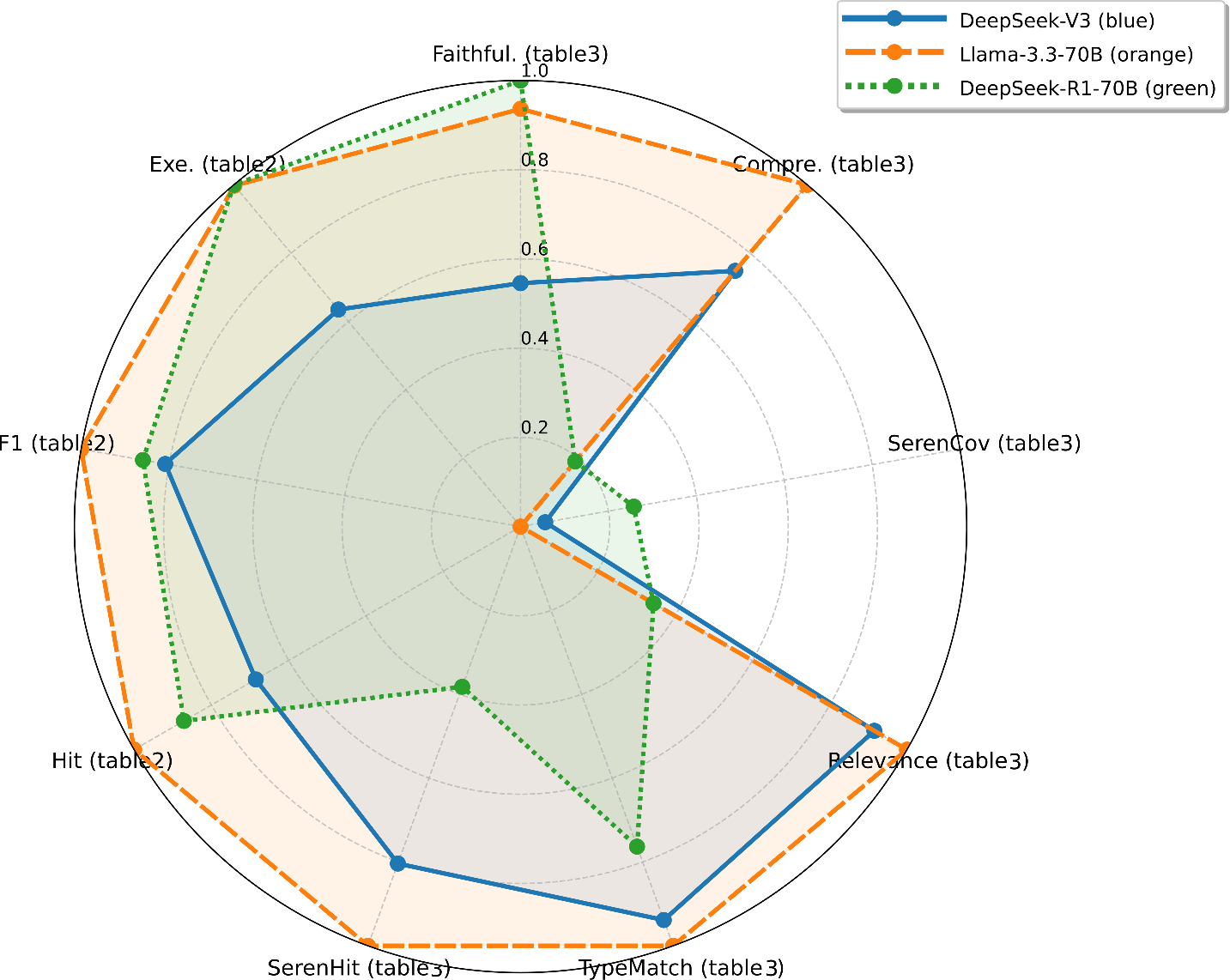}
    \centering
    \caption{Multi-step Performance Radar Chart}
    \label{fig:radar}
\end{figure*}

\stitle{Multi-Task Performance Compass. }
This radar chart shown in Fig.~\ref{fig:radar} clearly illustrates performance trade-offs across multiple tasks. DeepSeek-V3 excels at basic retrieval metrics (F1, Hit) but underperforms in Serendipity Coverage and SerenHit. In contrast, Llama-3-70B achieves high reasoning accuracy (Faithfulness and Comprehensiveness) yet only moderately captures serendipitous paths. DeepSeek-R1-70B demonstrates the opposite, effectively covering many serendipity paths but at the cost of reasoning accuracy. The absence of a dominant model across all metrics visually reinforces our earlier conclusion of \textit{no single winner}, suggesting the value of ensemble methods or Mixture-of-Experts (MoE) approaches.

\begin{figure*}[h!]
    \centering
    \includegraphics[width = \textwidth]{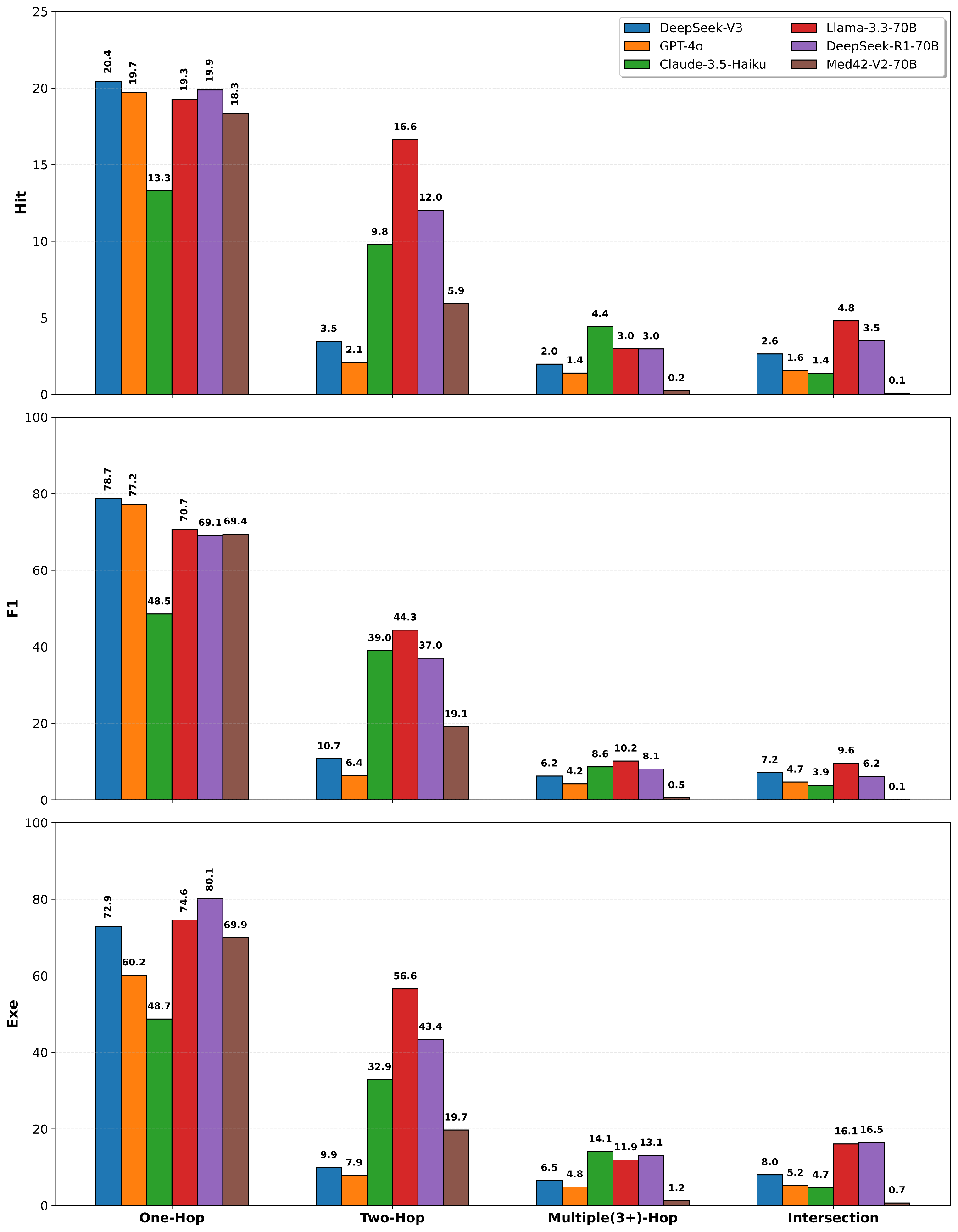}
    \centering
    \caption{Query Pattern vs. Retrieval Performance}
    \label{fig:bar}
\end{figure*}

\stitle{Query Pattern vs. Retrieval Performance.} As shown in Fig.~\ref{fig:bar}, model performance notably declines as query complexity increases. 
While all models achieve strong F1 and Hit scores on one-hop queries, 
results drop sharply for two-hop queries and especially for more complex queries ($\geq 3$-hop or intersection). 
This indicates that current LLMs, even the largest frontier models, still struggle significantly with complex multi-step reasoning and domain-specific context.

